\pdfoutput=1

\documentclass[11pt]{article}

\usepackage[preprint]{acl}

\usepackage{times}
\usepackage{latexsym}

\usepackage[T1]{fontenc}

\usepackage[utf8]{inputenc}

\usepackage{microtype}

\usepackage{inconsolata}

\usepackage{graphicx}

\usepackage{multicol}
\usepackage{multirow}
\usepackage{booktabs}
\usepackage{fvextra} 
\usepackage{amsmath}
\usepackage[breakable]{tcolorbox}
\usepackage{wrapfig}
\usepackage{varwidth}

\usepackage{hhline} 

\title{Map\&Make: Schema Guided Text to Table Generation}

\author{
\textsuperscript{*}Naman Ahuja \textsuperscript{1} \quad
\textsuperscript{*}Fenil Bardoliya\textsuperscript{1} \quad
Chitta Baral\textsuperscript{1} \quad
\textsuperscript{\dag}Vivek Gupta\textsuperscript{1} \\ 
\\ 
\textsuperscript{1}School of Computing and Augmented Intelligence, Arizona State University \\
\\
\texttt{nahuja11@asu.edu}, \texttt{fbardoli@asu.edu}, \texttt{chitta@asu.edu}, \texttt{\textsuperscript{\dag}vgupt140@asu.edu}
}

\begin{document}
\maketitle

\def\thefootnote{*}\footnotetext{These authors contributed equally to this work and share first authorship.}
\def\thefootnote{\dag}\footnotetext{This author supervised the research and serves as the corresponding author.}

\begin{abstract}
Transforming dense, detailed, unstructured text into an interpretable and summarised table, also colloquially known as Text-to-Table generation, is an essential task for information retrieval. Current methods, however, miss out on how and what complex information to extract; they also lack the ability to infer data from the text. In this paper, we introduce a versatile approach, Map\&Make, which ``dissects'' text into propositional atomic statements. This facilitates granular decomposition to extract the latent schema. The schema is then used to populate the tables that capture the qualitative nuances and the quantitative facts in the original text. Our approach is tested against two challenging datasets, Rotowire, renowned for complex and multi-table schema, and Livesum, which demands numerical aggregation. By carefully identifying and correcting hallucination errors in Rotowire, we aim to achieve a cleaner and more reliable benchmark. We evaluate our method rigorously on a comprehensive suite of comparative and referenceless metrics. Our findings demonstrate significant improvement results across both datasets with better interpretability in Text-to-Table generation. 
Moreover, through detailed ablation studies and analyses, we investigate the factors contributing to superior performance and validate the practicality of our framework in structured summarization tasks.
\footnote{Code and Data can be found here:\href {https://map-make.github.io}{https://map-make.github.io}}


\end{abstract}
\begin{figure}[h!]
    \centering
    \includegraphics[width=\linewidth]{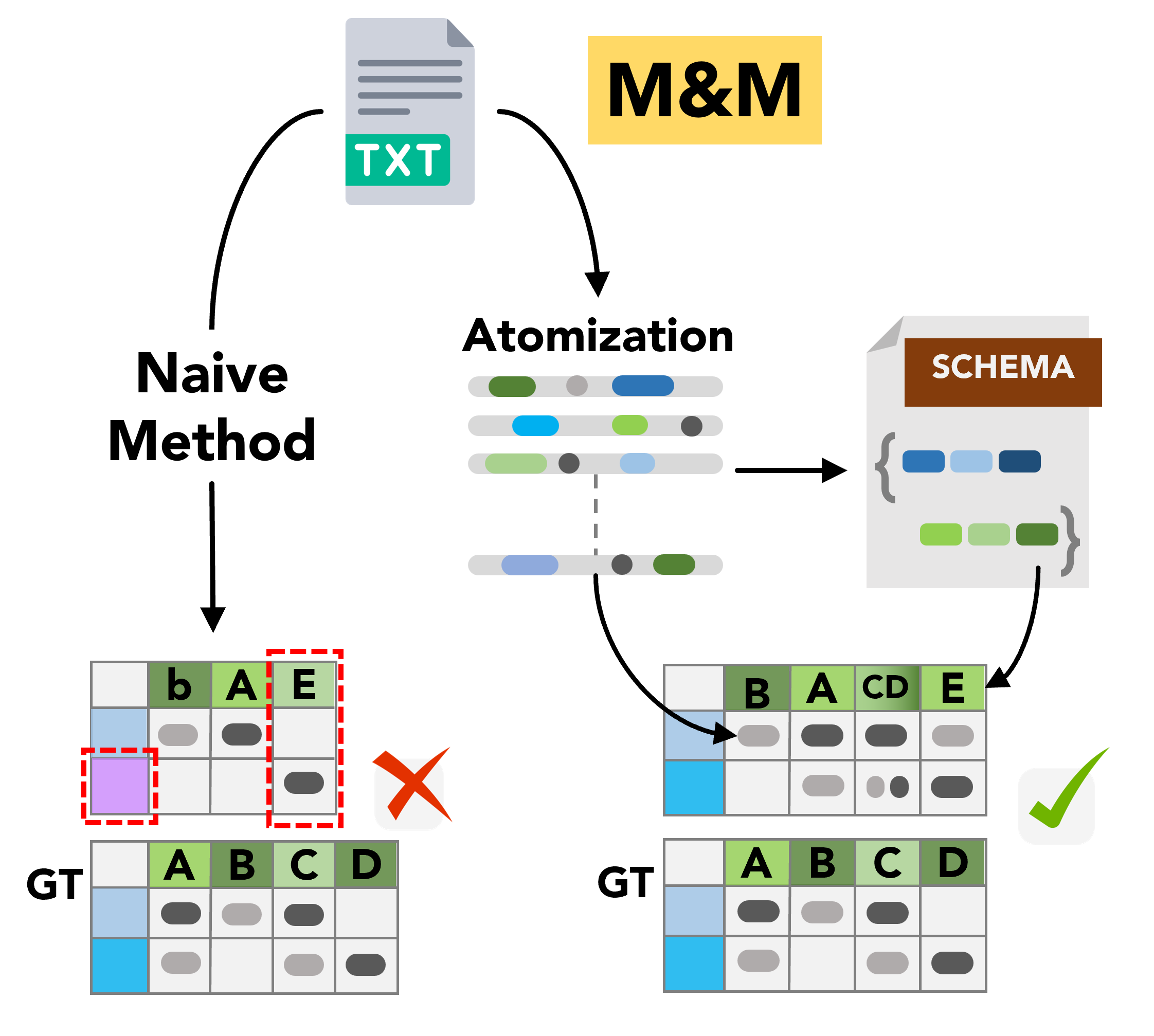}
    \caption{Comparison between Naive methods and our method for Text-to-Table generation.}
    \label{fig:abstract-mnm}
    \vspace{-1.0em}
\end{figure}

\section{Introduction}

Driven by the exceptional success of Large Language Models (LLMs) in tasks such as question answering~\cite{chen2020hybridqa,zhu2024tat}, text summarization~\cite{wiseman-etal-2017-challenges,wang2020cord}, and text data mining~\cite{li2023table,sui2024table}, recent works have shifted to explore the structured summarization of text. Reading lengthy texts is time-consuming, making it challenging to extract key information efficiently. Tables, as a widely used structured format, can organize data in a clear and interpretable manner, facilitating information retrieval and comprehension.
Early works~\cite{text-2-table-2022-acl,li2023sequence} treat text-to-table generation as an information extraction problem, viewing the table and its corresponding text representation as essentially similar. The pioneering work by~\cite{t3} extends beyond simple extraction, integrating and categorizing information in complex scenarios. However, to the best of our knowledge, all existing studies perform this task under a predefined schema, where information about the schema is either provided during fine-tuning~\cite{li2023sequence,text-2-table-2022-acl} or included in prompts for few-shot methods~\cite{t3,tang-etal-2024-struc}. This reliance on predefined schemas limits adaptability to open-domain text, where tabular structures may need to be inferred dynamically.

For more generalized systems, a hybrid approach is needed: leveraging schema induction techniques to recognize patterns in structured representations while retaining the flexibility to handle novel text narratives. This ensures robustness in managing both expected and unseen table structures. Nevertheless, tabular summarization of dense information without any prior knowledge of the underlying structure remains largely unexplored.

Regarding methodology, there are contrasting findings on the effect of fine-tuning LLMs versus zero-shot or few-shot prompting strategies across different datasets. Extensive benchmarks indicate that LLMs often perform suboptimally in zero-shot settings, missing relevant information and introducing unattested data or hallucinations. \cite{t3}

More sophisticated prompting techniques have been proposed to mitigate these issues~\cite{cot,khot2022decomposed}. In-context learning~\cite{brown2020language}, combined with chain-of-thought (CoT) prompting~\cite{cot} offers some improvements but can overfit to the provided shots~\cite{perez2021true}, reducing effectiveness on unseen data. Some studies~\cite{tang-etal-2024-struc,sundar2024gtbls} report performance gains from fine-tuning, whereas others~\cite{t3} observe no or negative improvements for table generation tasks.

Hence, there is a need for a generalized, schema-agnostic tabular summarization framework that can effectively \textit{"mine"} the underlying structure from free-form text and produce comprehensive information under a given instruction. 
In this work, we aim to address the aforementioned challenges with the following contributions:
\begin{enumerate}
    \item We introduce a generalized notion of Structured Summarization, evaluating LLMs’ capabilities on planning table schemas in Zero-Shot and One-Shot settings to summarize information under a given instruction exhaustively.
    \item We propose a generalizable framework, \textit{Map\&Make (M\&M)}, applicable to any Instruction-Driven Tabular Summarization task extending beyond simple extraction tasks.
    \item We present a manually corrected version of the Rotowire Benchmark, a popular text-to-table benchmark. This correction addresses hallucination and fidelity issues in previous versions, ensuring a fair evaluation setting for our methods.
    \item We conduct comprehensive experiments on multiple closed-source and open-source state-of-the-art LLMs on a diverse suite of metrics encompassing both table quality and information coverage. Results show that M\&M outperforms existing methods and exhibits robust generalization across different Tabular Summarization paradigms. Additionally, M\&M maintains stable performance on large text corpora, where other methods often degrade.

\end{enumerate}




\section {Map\&Make Framework} 
\begin{figure*}[ht!]
\centering
\includegraphics[width=\textwidth]{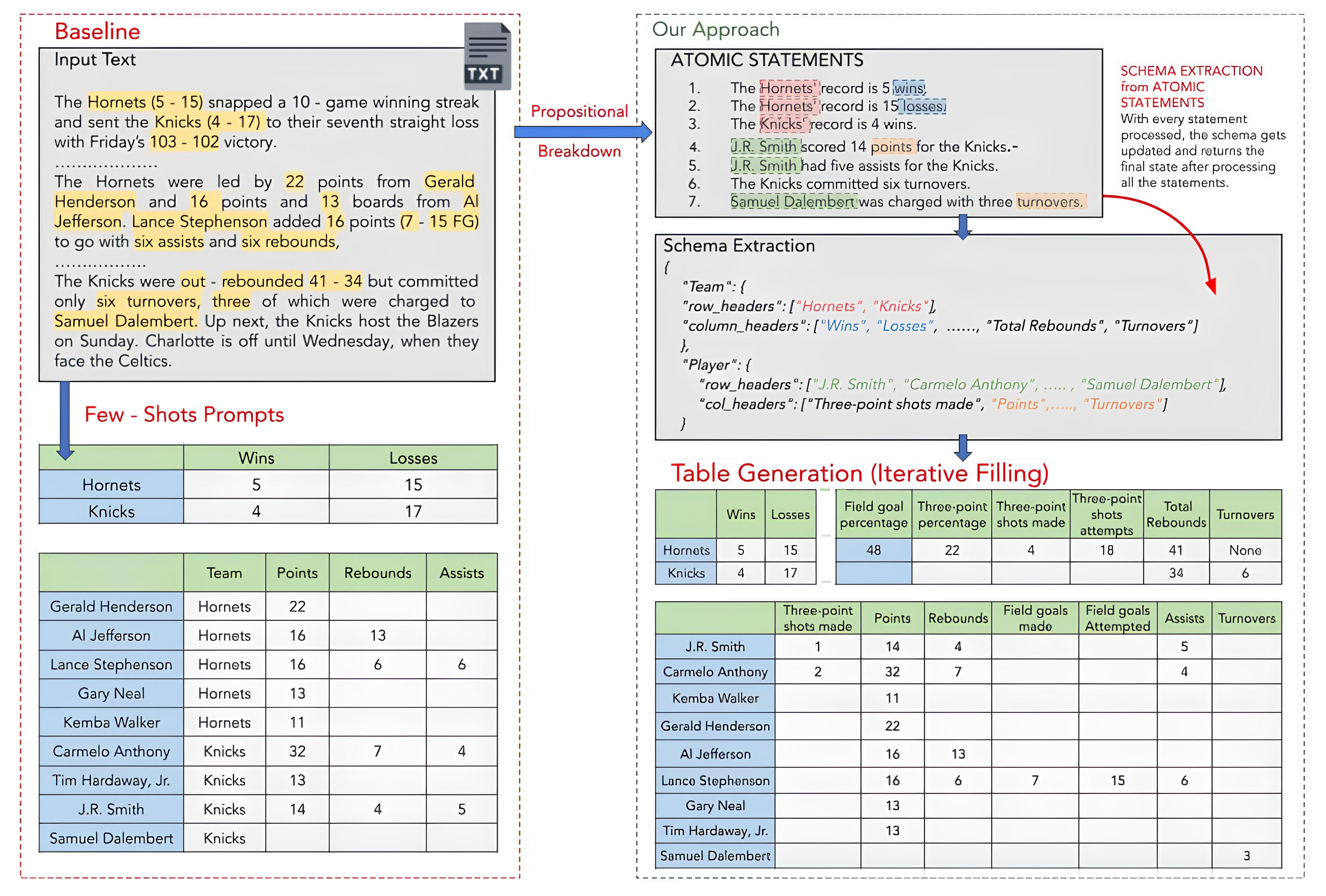}
\caption{An illustration of our approach. Propositional Breakdown segments the text to generate atomic statements b. Schema Extraction extracts the table structures to generate table schemas. c. Table Generation iteratively fills tables based on the atomic statements.}
\label{fig:method_illustration}
\vspace{-1.5em}
\end{figure*}
We propose a three-staged methodology applicable to any instruction-driven summarization task. Figure \ref{fig:method_illustration} illustrates our approach. Mirroring how humans construct tables—identifying key information, planning the layout, and filling in values—we employ a multi-agent prompting framework to enhance coverage and correctness.

\subsection{Propositional Atomization}
Propositional Atomization is the process of extracting atomic, self-contained facts from contextually embedded sentences. The benefits of this approach have been demonstrated in large-scale datasets and practical applications, such as in natural language inference and summary hallucination detection.\cite{Chen2022PropSegmEntAL} \cite{maynez-etal-2020-faithfulness}. We design a prompt to transform input statements into \textbf{atomic statements} adhering to five key properties: \textbf{Well-formedness}, ensuring grammatical correctness and adherence to linguistic conventions; \textbf{Atomicity}, maintaining the smallest meaningful semantic unit; \textbf{Self-containedness (Decontextualization)}, making each proposition independently understandable without requiring external context; \textbf{Support}, ensuring all propositions are explicitly derived from the input text; and \textbf{Comprehensiveness}, collectively covering all latent claims and facts from the original text. This targeted parsing resolves ambiguous entity attributions and prevents the conflation of overlapping or nested entities. The exact prompt is given in Appendix~\ref{sec:rotowire-atomic}. 

\subsection{Schema Extraction}
This part builds on propositional segmentation to extract tabular layouts. Directly planning schema results in incomplete coverage for inputs that require big tables to represent information. Hence, we iteratively build the table schema. The table schema is initialized as an empty list of row and column headers for each table. These lists are iteratively populated based on each atomic statement processed (as shown in Figure~\ref{fig:method_illustration}. Entities are mapped to row headers while their attributes are added to column headers. This iterative approach allows headers to evolve dynamically, enabling adaptation to the changing structure of incoming data, and improving coverage. The exact prompt is given in Appendix~\ref{sec:rotowire-schema}. 

\subsection{Table Generation}
\vspace{-0.5em}
In the final step, we perform table-filling on empty tables defined by the extracted schema in the previous step. We again, process every statement iteratively to update or fill cell values as per the summarization instruction. For example, for counting events from live text as in the LIVESUM benchmark, cell values keep increasing with every statement processed. Contrastively, for static text such as in Rotowire, cell values are fixed once. We explicitly instruct LLMs to return statement-wise updates for every cell value to ensure transparency and avoid hallucinations. Refer Appendix~\ref{sec:rotowire-table} for the exact prompt. 

\section{Benchmarks}
\vspace{-0.3em}
Pre-LLM methods for text-to-table generation~\cite{text-2-table-2022-acl}, \cite{li2023sequence}, \cite{pietruszka2024stable} have utilized four predominant benchmarks; Rotowire~\cite{wiseman-etal-2017-challenges}, E2E~\cite{novikova-etal-2017-e2e}, Wikibio~\cite{lebret-etal-2016-neural}, WikiTableText~\cite{bao2018table}. Initially proposed for \textbf{table-to-text} generation tasks, a majority of these benchmarks lack the structural complexity required to comprehensively evaluate state-of-the-art LLMs for information coverage in tabular summarization. In fact, E2E, WikiBio, and WikiTableText consist of only single tables with only two columns for every sample. Hence, we evaluate our framework on the repurposed version of the Rotowire proposed by \cite{text-2-table-2022-acl} and the recently released benchmark, Livesum~\cite{t3}. \\

Rotowire comprises dense post-match summaries of NBA games (2014–2017), where the task is to extract performance statistics to generate Player and Team Tables. In contrast, Livesum consists of live football commentary\footnote{https://www.bbc.com/sport/football}, requiring the aggregation of various events into team summary tables. The dataset also assigns a difficulty level: easy, medium, or hard to each column based on varying descriptions of events that influence how challenging inferring the right value is. Each dataset presents distinct challenges: Rotowire’s long text and multi-table structure make schema and sparse value extraction difficult, while Livesum demands identifying and aggregating events described in varying forms across the text. Together, these datasets offer a comprehensive testbed for evaluating text-to-table generation frameworks. More details are provided in Appendix~\ref{sec:Livesum Characteristics}

\subsection{Corrections and Fidelity Issues:}

In our initial explorations, we observed several discrepancies between the text and the ground truth tables in the data presented by \cite{text-2-table-2022-acl}. These discrepancies propagate on other benchmarks developed on this data \cite{tang-etal-2024-struc} inhibiting fair and reliable evaluations of developed frameworks. We suspect these challenges stem from LLM-driven attribution methods,  and existing problems in validating information \cite{adewumi2024limitations}, \cite{yue2023automatic},\cite{patel-etal-2024-towards}. Thus the authors manually validate every sample in the test set. Keeping ground truth text the same, we attribute every row, column, and cell value to the text and update tables. 

\begin{table}[h]
\vspace{-0.75em}
\centering
\small
\begin{tabular}{lcccccc}
\hline \hline
\multicolumn{1}{c}{\multirow{2}{*}{\textbf{Table}}} &
  \multicolumn{2}{c}{\textit{\textbf{Cell}}} &
  \multicolumn{2}{c}{\textit{\textbf{Row}}} &
  \multicolumn{2}{c}{\textit{\textbf{Col}}} \\ \cline{2-7} 
\multicolumn{1}{c}{} &
  \textbf{H} &
  \textbf{MI} &
  \textbf{H} &
  \textbf{MI} &
  \textbf{H} &
  \textbf{MI} \\ \hline
\textbf{}       & \multicolumn{6}{c}{\textbf{Original to Strucbench}}  \\ \hline
\textbf{Team}   & 1219     & 1271    & 8      & 8     & 627    & 626   \\
\textbf{Player} & 1390     & 1270    & 68     & 62    & 135    & 129   \\ \hline
                & \multicolumn{6}{c}{\textbf{Original to Corrected}}   \\ \hline
\textbf{Team}   & 613      & 1137    & 21     & 50    & 329    & 528   \\
\textbf{Player} & 7310     & 1752    & 85     & 82    & 1000   & 188   \\ \hline
                & \multicolumn{6}{c}{\textbf{Strucbench to Corrected}} \\ \hline
\textbf{Team}   & 721      & 1247    & 21     & 50    & 385    & 585   \\
\textbf{Player} & 8104     & 2666    & 140    & 143   & 1077   & 271   \\ \hline \hline
\end{tabular}%
\caption{Total counts of corrected rows, columns, and cells across error types for Rotowire. \textbf{H} denotes Hallucination, and \textbf{MI} represents Missing Information. Here, a row or a column is flagged as hallucinated/missing if it contains at least one erroneous entry.}

\label{tab:dataset-correction}
\vspace{-0.75em}
\end{table}

Our findings are reported in Table~\ref{tab:dataset-correction}, where we compare our corrected version with the Original~\cite{text-2-table-2022-acl}, and STRUCBENCH~\cite{tang-etal-2024-struc} benchmark which is another version of Rotowire. More details about the correction strategy can be found in Appendix~\ref{sec:rotowire-corrections}

\section{Experimental Setup}
We describe the different prompting techniques, LLMs, and evaluation metrics employed in this work.

\subsection{Models and Baselines}
We experiment with M\&M on both closed-source and open-source LLMs. Among proprietary models, we use Gemini 2.0 Flash experimental and GPT-4o~\cite{gpt4o} (\textit{gpt-4o-2024-08-06}), maintaining consistent configurations for \textit{temperature}, \textit{top\_p}, and \textit{top\_k} across experiments. Additionally, we assess performance on open-source models such as Llama 3.3-70B Instruct.
\footnote{https://github.com/meta-llama/llama-models/blob/main/models/llama3\_3/LICENSE}. 

For baselines, we employ various prompting strategies. \textbf{Zero-Shot CoT}~\cite{autocot} directly organizes and populates tables by leveraging step-by-step reasoning, while \textbf{One-Shot CoT}~\cite{cot} enhances this by incorporating a single example in the prompt. We also compare against \textbf{Text-Tuple-Table (T\textsuperscript{3})}~\cite{t3}, which extracts structured tuples (\textit{subject-object-verb} or \textit{subject-attribute-value}) before table generation. Originally schema-dependent, we adapt T\textsuperscript{3} into a schema-agnostic setting, alongside its unified variant \textbf{T\textsuperscript{3}D}, which integrates tuple extraction and table construction in a single prompt. To evaluate the efficacy of our framework, we experiment with two variants of our approach; \textbf{M\&M - 3S}, performs each task sequentially, where as \textbf{M\&M - U}; a unified variant of our approach, performs all the tasks in a single LLM call. We face several challenges on testing M\&M on the Livesum dataset on smaller models due to the large input sizes, resulting in inefficient iterative table filling and same statements being duplicated multiple times . This aligns with the findings of the benchmark authors \cite{t3}, where they face similar challenges in building prompting pipelines for this dataset over smaller models. Moreover, the authors also show the no/negative improvements from fine-tuning on this dataset. Hence, we only experiment with prompting-based approaches for this problem.

\subsection{Evaluation Metrics}

Due to the high variance in table structures, where the same information can be presented in different formats, we utilize a diverse set of both LLM-based and Non-LLM-based metrics to ensure a comprehensive evaluation of correctness and competeness. \\

\noindent \textbf{String Similarity Metrics:} We utilize Exact Match (EM), CHRF~\cite{popovic-2015-chrf}, and BERTScore~\cite{Zhang*2020BERTScore:} as proposed by~\cite{text-2-table-2022-acl} for reference-based table evaluation. To measure information coverage, each (row, column, cell) tuple from the ground truth table is mapped to its most similar counterpart in the generated table, assigning the highest similarity score to that tuple. Rows Headers and Column Headers are evaluated in a similar fashion. The final EM, CHRF, and BERT scores are averaged over all tuples, as reported in Table \ref{tab:rotowire-non-llm}. 
For Livesum, where the correctness of generated outputs is determined by the model counting the correct number of occurrences of an event, i.e numbers, metrics like CHRF and BERTScore are unsuitable. Instead, we use Root Mean Squared Error (RMSE) defined as:

\begin{small}
\[
RMSE = \sqrt{\frac{\sum_{i=1}^{n} (y_i - \tilde{y}_i)^2}{n}}
\]
\end{small}
where \( n \) denotes the total number of cells, and \( y_i \) and \( \tilde{y}_i \) represent the content of the cell at index \( i \) in the ground truth table and the generated table, respectively. For 2D tables, the index \( i \) is determined by flattening the table into a 1D sequence. Additionaly, for exact cell value matching, we report Error Rate (ER \%) as a percentage of erroneous cells.
\\


\noindent \textbf{Specialized Metrics:} Evaluating table cells (or tuples) using similarity-based metrics independently without considering contextual information from the neighbouring cells can lead to incorrect penalization of good tables, or incorrect rewarding of bad tables. TabEval~\cite{ramu-etal-2024-bad} addresses these challenges by calculating table similarity by \textit{unrolling} tables into a list of atomic statements and computes the entailment between statements generated from output tables and their ground truths. Additionally, we also utilize Auto-QA as an information coverage metric~\cite{jain-etal-2024-structsum} defined as:

\begin{small}
\begin{equation}
    \text{Cov}(\mathcal{T}) = \frac{\sum_{i=1}^{|G(\mathcal{S})|} E_{(q_i, a_i)} \left[ Q(\{\mathcal{T}_j\}, q_i) \right]}{|G(\mathcal{S})|}
\end{equation}
\end{small}

where \( G(\mathcal{S}) \) represents a set of Question-Answer pairs \( (q_i, a_i) \) generated by an LLM from the input text \( \mathcal{S} \). The function \( Q(\{\mathcal{T}_j\}, q) \) denotes the LLM’s response to question \( q \), derived from a set of generated tables \( \{\mathcal{T}_j\} \). The term \( E_{(q, a)}[x] \) evaluates whether the LLM’s response \( x \) aligns with the reference answer \( a \) for the given question \( q \). We report accuracy as the percentage of correctly answered questions based on the available tables.

\section{Results and Discussion}


\begin{table}[ht]
\small
\setlength{\tabcolsep}{2.5pt}
\centering
\resizebox{\linewidth}{!}{%
\begin{tabular}{lccccccccc}
\hline \hline
\multicolumn{1}{c}{\multirow{2}{*}{\textbf{Method}}} &
  \multicolumn{3}{c}{\textit{\textbf{EM}}} &
  \multicolumn{3}{c}{\textit{\textbf{CHRF}}} &
  \multicolumn{3}{c}{\textit{\textbf{BERT}}} \\ \cline{2-10} 
\multicolumn{1}{c}{} &
  \textbf{Cell} &
  \textbf{Row} &
  \textbf{Col} &
  \textbf{Cell} &
  \textbf{Row} &
  \textbf{Col} &
  \textbf{Cell} &
  \textbf{Row} &
  \textbf{Col} \\ \hline
\multicolumn{1}{c}{\textit{\textbf{}}} &
  \multicolumn{9}{c}{\textit{\textbf{GPT-4o Zero-Shot}}} \\ \hline
\textbf{CoT} &
  \textbf{20.75} &
  \textbf{51.58} &
  30.13 &
  39.62 &
  \textbf{81.46} &
  49.05 &
  38.25 &
  62.31 &
  59.59 \\
\textbf{T\textsuperscript{3} - D} &
  18.39 &
  51.32 &
  29.01 &
  37.13 &
  81.24 &
  46.37 &
  36.44 &
  62.35 &
  58.25 \\
\textbf{M\&M - U} &
  19.95 &
  50.43 &
  \textbf{30.39} &
  \textbf{41.99} &
  78.97 &
  \textbf{53.12} &
  \textbf{42.90} &
  \textbf{62.57} &
  \textbf{64.82} \\ \hline
\multicolumn{1}{c}{\textit{\textbf{}}} &
  \multicolumn{9}{c}{\textit{\textbf{Gemini-2.0 Zero-Shot}}} \\ \hline
\textbf{CoT} &
  \textbf{21.90} &
  51.85 &
  29.20 &
  36.65 &
  80.69 &
  44.23 &
  38.58 &
  63.78 &
  55.76 \\
\textbf{T\textsuperscript{3} - D} &
  20.77 &
  \textbf{52.92} &
  22.83 &
  32.38 &
  \textbf{81.90} &
  38.80 &
  37.54 &
  \textbf{64.65} &
  54.05 \\
\textbf{M\&M - U} &
  19.76 &
  48.05 &
  \textbf{31.51} &
  \textbf{42.61} &
  75.00 &
  \textbf{53.19} &
  \textbf{41.22} &
  60.63 &
  \textbf{60.65} \\ \hline
\multicolumn{1}{c}{\textit{\textbf{}}} &
  \multicolumn{9}{c}{\textit{\textbf{Llama-3.3 70B Zero-Shot}}} \\ \hline
\textbf{CoT} &
   \textbf{21.44} &
   51.42 &
   39.37 &
   40.79 &
   80.59 &
   51.40 &
   36.70 &
   \textbf{63.45} &
   55.73 \\
\textbf{T\textsuperscript{3} - D} &
   21.05 &
   51.57 &
   \textbf{39.40} &
   \textbf{42.04} &
   \textbf{80.77} &
   \textbf{53.36} &
   \textbf{37.31} &
   62.93 &
   \textbf{57.49} \\
\textbf{M\&M - U} &
  19.26 &
  \textbf{51.64} &
  27.52 &
  35.63 &
  79.86 &
  45.45 &
  36.95 &
  63.28 &
  55.30 \\ \hline
\multicolumn{1}{c}{\textit{\textbf{}}} &
  \multicolumn{9}{c}{\textit{\textbf{GPT-4o One-Shot}}} \\ \hline
\textbf{CoT} &
  33.30 &
  88.08 &
  35.96 &
  45.42 &
  91.98 &
  50.60 &
  57.11 &
  91.18 &
  60.28 \\
\textbf{T\textsuperscript{3} - D} &
  34.53 &
  85.98 &
  38.36 &
  47.96 &
  91.35 &
  53.31 &
  57.83 &
  89.79 &
  61.90 \\
\textbf{T\textsuperscript{3}} &
  17.92 &
  53.92 &
  34.43 &
  40.94 &
  82.69 &
  52.25 &
  39.21 &
  64.49 &
  61.77 \\
\textbf{M\&M - U} &
  \textbf{51.20} &
  \textbf{90.87} &
  \textbf{55.25} &
  \textbf{68.27} &
  \textbf{93.61} &
  \textbf{73.38} &
  \textbf{74.39} &
  \textbf{95.24} &
  \textbf{77.88} \\
\textbf{M\&M - 3S} &
  35.24 &
  64.06 &
  52.72 &
  61.89 &
  85.06 &
  72.49 &
  57.90 &
  73.30 &
  76.80 \\ \hline
\multicolumn{1}{c}{\textit{\textbf{}}} &
  \multicolumn{9}{c}{\textit{\textbf{Gemini-2.0 One-Shot}}} \\ \hline
\textbf{CoT} &
  40.65 &
  \textbf{91.57} &
  43.21 &
  55.19 &
  \textbf{93.92} &
  60.10 &
  64.60 &
  \textbf{95.53} &
  66.22 \\
\textbf{T\textsuperscript{3} - D} &
  38.89 &
  87.84 &
  42.91 &
  55.19 &
  92.90 &
  60.70 &
  63.31 &
  92.54 &
  67.18 \\
\textbf{T\textsuperscript{3}} &
  28.61 &
  65.89 &
  40.21 &
  53.38 &
  85.94 &
  62.05 &
  51.91 &
  74.19 &
  69.09 \\
\textbf{M\&M - U} &
  56.44 &
  90.77 &
  60.82 &
  71.38 &
  93.50 &
  76.34 &
  76.12 &
  94.85 &
  78.37 \\
\textbf{M\&M - 3S} &
  \textbf{61.78} &
  89.49 &
  \textbf{67.37} &
  \textbf{76.12} &
  92.92 &
  \textbf{81.57} &
  \textbf{79.48} &
  93.86 &
  \textbf{82.69} \\ \hline 
\multicolumn{1}{c}{\textit{\textbf{}}} &
  \multicolumn{9}{c}{\textit{\textbf{Llama-3.3 70B One-Shot}}} \\ \hline
\textbf{CoT} &
  39.57 &
  91.82 &
  41.52 &
  53.08 &
  93.77 &
  57.41 &
  62.69 &
  94.61 &
  64.09 \\
\textbf{T\textsuperscript{3} - D} &
  38.02 &
  89.38 &
  41.61 &
  50.44 &
  91.79 &
  56.87 &
  61.10 &
  92.09 &
  63.23 \\
\textbf{T\textsuperscript{3}} &
  33.82 &
  85.41 &
  38.84 &
  49.75 &
  89.95 &
  57.17 &
  59.04 &
  89.52 &
  64.54 \\
\textbf{M\&M - U} &
  \textbf{44.27} &
  \textbf{92.78} &
  \textbf{47.27} &
  \textbf{60.54} &
  \textbf{94.85} &
  \textbf{65.62} &
  \textbf{66.96} &
  \textbf{95.77} &
  \textbf{70.25} \\
\textbf{M\&M - 3S} &
  41.01 &
  72.02 &
  44.94 &
  54.47 &
  74.39 &
  58.87 &
  58.15 &
  74.99 &
  61.42 \\ \hline \hline
\end{tabular}%
}
\caption{\small Performance comparison of different methods on Rotowire across models using various string similarity metrics}
\label{tab:rotowire-non-llm}
\end{table}

\begin{table}[h]
\small
\centering
\setlength{\tabcolsep}{2.5pt}
\resizebox{\linewidth}{!}{%
\begin{tabular}{lccccccc}
\hline \hline
\multicolumn{1}{c}{\textbf{}} &
  \multicolumn{6}{c}{\textbf{TabEval}} &
  \textbf{Auto-QA} \\ \cline{2-8} 
\multicolumn{1}{c}{\textbf{Method}} &
  \multicolumn{2}{c}{\textit{\textbf{Correctness}}} &
  \multicolumn{2}{c}{\textit{\textbf{Completeness}}} &
  \multicolumn{2}{c}{\textit{\textbf{Overall}}} &
  \multirow{2}{*}{\textit{\textbf{Accuracy}}} \\ \cline{2-7}
 &
  \textbf{Team} &
  \textbf{Player} &
  \textbf{Team} &
  \textbf{Player} &
  \textbf{Team} &
  \textbf{Player} &
   \\ \hline
\multicolumn{8}{c}{\textit{\textbf{GPT-4o Zero-Shot}}} \\ \hline
\textbf{CoT} &
  57.32 &
  62.78 &
  39.62 &
  79.92 &
  42.66 &
  67.64 &
  70.13 \\
\textbf{T\textsuperscript{3} - D} &
  \textbf{58.60} &
  \textbf{73.85} &
  \textbf{42.44} &
  79.37 &
  \textbf{45.36} &
  \textbf{73.66} &
  66.69 \\
\textbf{M\&M - U} &
  43.74 &
  68.40 &
  41.74 &
  \textbf{85.96} &
  38.27 &
  73.20 &
  \textbf{76.95} \\ \hline
\multicolumn{8}{c}{\textit{\textbf{Gemini-2.0 Zero-Shot}}} \\ \hline
\textbf{CoT} &
  \textbf{73.41} &
  66.94 &
  36.27 &
  \textbf{81.35} &
  \textbf{44.72} &
  \textbf{71.39} &
  60.52 \\
\textbf{T\textsuperscript{3} - D} &
  55.82 &
  63.70 &
  31.57 &
  80.80 &
  36.65 &
  68.62 &
  61.94 \\
\textbf{M\&M - U} &
  58.01 &
  \textbf{68.93} &
  \textbf{44.19} &
  77.94 &
  46.53 &
  68.72 &
  \textbf{64.63} \\ \hline
\multicolumn{8}{c}{\textit{\textbf{Llama-3.3 70B Zero-Shot}}} \\ \hline
\textbf{CoT} &
  \textbf{67.63} &
  70.57 &
  30.66 &
  13.35 &
  38.09 &
  16.21 &
  54.24 \\
\textbf{T\textsuperscript{3} - D} &
  63.60 &
  \textbf{74.88} &
  \textbf{33.43} &
  \textbf{34.23} &
  \textbf{40.56} &
  \textbf{33.14} &
  52.41 \\
\textbf{M\&M - U} &
  61.15 &
  72.32 &
  27.78 &
  14.61 &
  33.55 &
  17.88 &
  \textbf{59.17} \\ \hline
\multicolumn{8}{c}{\textit{\textbf{GPT-4o One-Shot}}} \\ \hline
\textbf{CoT} &
  80.23 &
  86.74 &
  56.26 &
  57.74 &
  64.34 &
  65.57 &
  41.42 \\
\textbf{T\textsuperscript{3} - D} &
  80.24 &
  \textbf{87.56} &
  56.12 &
  64.34 &
  64.09 &
  70.84 &
  38.08 \\
\textbf{T\textsuperscript{3}} &
  \textbf{81.65} &
  85.39 &
  46.43 &
  60.66 &
  56.05 &
  66.61 &
  39.46 \\
\textbf{M\&M - U} &
  66.28 &
  86.16 &
  \textbf{77.65} &
  \textbf{88.15} &
  \textbf{69.56} &
  \textbf{85.85} &
  69.51 \\
\textbf{M\&M - 3S} &
  48.78 &
  74.22 &
  61.43 &
  87.34 &
  51.01 &
  78.56 &
  \textbf{80.38} \\ \hline
\multicolumn{8}{c}{\textit{\textbf{Gemini-2.0 One-Shot}}} \\ \hline
\textbf{CoT} &
  \textbf{82.91} &
  \textbf{91.51} &
  63.50 &
  75.89 &
  69.72 &
  81.06 &
  60.52 \\
\textbf{T\textsuperscript{3} - D} &
  81.86 &
  91.08 &
  65.18 &
  76.27 &
  70.01 &
  80.92 &
  47.94 \\
\textbf{T\textsuperscript{3}} &
  75.70 &
  87.22 &
  61.75 &
  82.07 &
  64.93 &
  82.68 &
  56.58 \\
\textbf{M\&M - U} &
  75.27 &
  91.17 &
  77.47 &
  92.92 &
  \textbf{74.80} &
  \textbf{91.36} &
  64.88 \\
\textbf{M\&M - 3S} &
  65.89 &
  83.54 &
  \textbf{79.49} &
  \textbf{92.43} &
  69.95 &
  86.71 &
  \textbf{71.21} \\ \hline 
  \multicolumn{8}{c}{\textit{\textbf{Llama-3.3 70B One-Shot}}} \\ \hline
\textbf{CoT} &
  \textbf{80.93} &
  80.41 &
  59.33 &
  \textbf{47.79} &
  66.33 &
  \textbf{47.29} &
  54.20 \\
\textbf{T\textsuperscript{3} - D} &
  69.54 &
  \textbf{80.86} &
  48.57 &
  45.32 &
  55.21 &
  45.92 &
  46.57 \\
\textbf{T\textsuperscript{3}} &
  65.31 &
  70.66 &
  64.38 &
  36.38 &
  61.13 &
  33.03 &
  53.49 \\
\textbf{M\&M - U} &
  74.33 &
  76.47 &
  \textbf{71.87} &
  43.34 &
  \textbf{70.57} &
  39.53 &
  55.96 \\
\textbf{M\&M - 3S} &
  47.92 &
  57.12 &
  53.59 &
  34.04 &
  48.21 &
  31.34 &
  \textbf{64.61} \\ \hline \hline
\end{tabular}%
}
\caption{\small Performance comparison using TabEval and AutoQA on Rotowire across strategies on various models.}
\label{tab:rotowire-llm-based}
\vspace{-1.0em}
\end{table}

Our experiments demonstrate Map\&Make outperforms other prompting strategies improving information coverage and correctness significantly across all models and evaluation metrics, showing excellent generalization capabilities.  

\subsection{Performance on Rotowire} From Table~\ref{tab:rotowire-non-llm}, we observe that M\&M outperforms other baselines such as CoT, T\textsuperscript{3} consistently in the cell- and column-level metrics. In a zero-shot setting, M\&M improves coverage by up to 32\% (from CHRF) at the column level and 32\% at the cell level. In a one-shot setting, M\&M improves by about 29\% at the cell level and by 27\% at the column level.  We see minimal to no improvement in the Row-level metrics across all metrics for both zero-shot and one-shot settings.  The row headers in all the tables consist of proper nouns (Player Names, Team Names), and LLMs perform well in extracting named entities~\cite{villena2024llmner}. At the table level, we observe up to 42\% improvement in TabEval scores (Table~\ref{tab:rotowire-llm-based}) for the Team tables and 22\% for the Player table. A bigger improvement is seen in the Team Tables as the global set of column headers for team tables is larger, introducing more variety in the team schemas. Also, we observe a decrease in the Tab-Eval Correctness. This is discussed in detail in the Appendix~\ref{sec:Why precision decreases}, as the Ground Truth tables in the RotoWire dataset consist only of statistical values, and M\&M being a loosely supervised method also extracts non-statistical columns ( such as Injury Status and Average Player Performance, etc.), attributes not present in the Ground Truth, leading to decreased correctness scores. We validate the precision of our generated tables using the Auto-QA metric and observe an improvement of up to 15\% in coverage from CoT baselines.
\begin{table}[h]
\small
\centering
\setlength{\tabcolsep}{3.5pt}
\resizebox{\linewidth}{!}{%
\begin{tabular}{lcccccccc}
\hline \hline
\multicolumn{1}{c}{\multirow{2}{*}{\textbf{Method}}} & 
\multicolumn{2}{c}{\textbf{Easy}} & 
\multicolumn{2}{c}{\textbf{Medium}} & 
\multicolumn{2}{c}{\textbf{Hard}} & 
\multicolumn{2}{c}{\textbf{Average}} \\ \cline{2-9} 
\multicolumn{1}{c}{} & \textbf{RMSE} & \textbf{ER} & \textbf{RMSE} & \textbf{ER} & \textbf{RMSE} & \textbf{ER} & \textbf{RMSE} & \textbf{ER} \\ \hline
\multicolumn{1}{c}{\textit{\textbf{}}} & \multicolumn{8}{c}{\textit{\textbf{GPT-4o Zero-Shot}}} \\ \hline
\textbf{CoT}                   &\textbf{0} & \textbf{0} & 1.55 & 47.71 & 1.84 & 79.84 & 1.55 & 48.14 \\
\textbf{T\textsuperscript{3}D}  & 0.05 & 1.82 & 1.93 & 53.66 & 2.76 & 84.76 & 2.04 & 50.10 \\
\textbf{M\&M - U} & 0.03 & 1.92 & \textbf{0.80} & \textbf{26.56} & \textbf{1.50} & \textbf{55.19} & \textbf{0.97} & \textbf{24.63} \\ \hline
\multicolumn{1}{c}{\textit{\textbf{}}} & \multicolumn{8}{c}{\textit{\textbf{Gemini-2.0 Zero-Shot}}} \\ \hline
\textbf{CoT}  & \textbf{0.05} & \textbf{2.51} & 2.00 & 60.78 & 2.82 & 87 & 2.09 & 53.92 \\
\textbf{T\textsuperscript{3}D} & 0.05 & 2.82 & 3.45 & 60.20 & 4.39 & 93.26 & 3.34 & 54.44 \\
\textbf{M\&M - U}  & 0.08 & 3.82 & \textbf{0.87} & \textbf{31.31} & \textbf{2.63} & \textbf{80.31} & \textbf{1.47} & \textbf{35.00} \\ \hline
\multicolumn{1}{c}{\textit{\textbf{}}} & \multicolumn{8}{c}{\textit{\textbf{GPT-4o One-Shot}}} \\ \hline
\textbf{CoT}  & 0.05 & 2.62 & 2.05 & 55.30 & 1.73 & 76.86 & 1.61 & 45.34 \\
\textbf{T\textsuperscript{3}D} & 0.08 & \textbf{0.43} & 1.46 & 46.62 & 2.73 & 84.73 & 1.75 & 44.91 \\
\textbf{T\textsuperscript{3}} & 0.23 & 16.5 & 1.48 & 34.16 & 2.39 & 51.91 & 1.76 & 35.62 \\
\textbf{M\&M - U} & \textbf{0.03} & 1.85 & 0.81 & 31.06 & 1.96 & 68.10 & 1.19 & 32.94 \\
\textbf{M\&M - 3S}  & 0.07 & 3.44 & \textbf{0.63} & \textbf{25.48} & \textbf{0.93} & \textbf{32.34} & \textbf{0.71} & \textbf{21.77} \\ \hline
\multicolumn{1}{c}{\textit{\textbf{}}} & \multicolumn{8}{c}{\textit{\textbf{Gemini-2.0 One-Shot}}} \\ \hline
\textbf{CoT}   & 0.01 & 0.63 & 1.69 & 58.31 & 2.23 & 82.80 & 1.68 & 48.32 \\
\textbf{T\textsuperscript{3}D} & \textbf{0.01} & \textbf{0.54} & 1.21 & 49.63 & 2.34 & 85.12 & 1.49 & 46.44 \\
\textbf{T\textsuperscript{3}} & 0.19 & 9.26 & 0.93 & 26.32 & 2.51 & 54.72 & 1.59 & 29.10 \\
\textbf{M\&M - U}  & 0.04 & 1.45 & 0.89 & 36.64 & 1.98 & 69.93 & 1.21 & 34.65 \\
\textbf{M\&M - 3S}  & 0.10 & 4.58 & \textbf{0.60} & \textbf{23.72} & \textbf{0.89} & \textbf{33.13} & \textbf{0.70} & \textbf{21.37} \\ \hline \hline
\end{tabular}%
}
\caption{\small Performance comparison using string similarity metrics across different categories showing Error Rates (in \%) and RMSE scores of GPT-4o and Gemini-2.0-flash-exp for Livesum}
\label{tab:livesum-main-table}
\vspace{-1.0em}
\end{table}

\subsection{Performance on Livesum} 
Table \ref{tab:livesum-main-table} compares the difficulty wise and overall performance of M\&M with other techniques. M\&M significantly enhances performance by reducing overall error rate by upto 35\% and RMSE by upto 29\% in zero shot setting, and by upto 55\% in error and 57\% in RMSE in one-shot setting. CoT also outperforms T\textsuperscript{3}-D in most experiments, signifying lack of generalization across a schema-agnostic setting. However we see consistent improvement across T\textsuperscript{3} from CoT baselines, as it uses code-generation to integrate, or count the events before generating the final table. On comparing performance of zero-shot - vs one-shot methods, we see our methods consistently improve performance across all methods and models.
This finding is consistent across both benchmarks, hence showcasing emergent capabilities as an adaptable one-shot framework.

\subsection{Performance Across LLMs:} In the Zero-Shot setting, GPT-4o and Gemini 2.0 Flash Exp demonstrate largely comparable performances across most metrics, with only minor variations. The largest observed difference is a 4.8-point delta in Column CHRF score for CoT prompting, where GPT slightly outperforms Gemini, and a 6-point delta in Column CHRF for M\&M prompting, favoring Gemini. In the One-Shot setting, Gemini exhibits a slight edge, with improvements of up to 4.8 in CHRF Column score for CoT prompting, and 2.1 in Cell-level coverage for M\&M Multistep. The overall TabEval score difference remains minimal, with a delta of just 0.2 in favor of Gemini. These results highlight the robustness and adaptability of our framework, maintaining stable and consistent performance across diverse datasets and evaluation metrics.


\begin{table}[ht]
\centering           
\small
\small
\begin{tabular}{lrr}
\hline\hline
 & \textbf{CoT} & \textbf{M\&M} \\
\hline
\multicolumn{3}{c}{\textbf{Extra Information}} \\
\hline
\textbf{Team Rows}       & \footnotesize 0.05 & \footnotesize 0 \\
\textbf{Player Rows}     & \footnotesize 0.41 & \footnotesize 0.06 \\
\textbf{Team Column}   & \footnotesize 0    & \footnotesize 0.09 \\
\textbf{Player Columns}  & \footnotesize 0.07 & \footnotesize 1.93 \\
\hline
\multicolumn{3}{c}{\textbf{Missing Information}} \\
\hline
\textbf{Team Rows}       & \footnotesize 0.019 & \footnotesize 0 \\
\textbf{Player Rows}     & \footnotesize 0.11  & \footnotesize 0.04 \\
\textbf{Team Columns}    & \footnotesize 1.04  & \footnotesize 0 \\
\textbf{Player Columns}  & \footnotesize 2.6   & \footnotesize 0.21 \\
\hline \hline
\end{tabular}
\caption{Average Per-Table Error Counts Across Rows and Columns for Rotowire}
\label{tab: rotowire-error analysis}
\vspace{-1.0em}
\end{table}


\subsection{Error Analysis}
To qualitatively analyze and pinpoint exactly where information loss occurs in table generation, we employ a table-transformation agent to transform CoT and M\&M generated outputs to exactly match the Row and Column headers of the Ground Truth tables. The agent first maps the given rows and columns of the inputs with the Ground Truth schema and subsequently populates the table. Rows and Columns that occur in the model outputs but are not present in the ground truth are reported separately (extra information) and vice versa (missing information). This agent facilitates a more granular analysis of each table giving insights into what information a strategy misses out on. 

\textbf{Rotowire}: As shown in Table \ref{tab: rotowire-error analysis}. 
Row-wise coverage improves from the CoT baseline when employing Map\&Make reducing missing rows from 0.05 to 0 per table for Teams, and from 0.41 to 0.06 per table for players. We also observe a substantial improvement in Column level coverage. CoT baselines miss out on covering columns both in Team and Player tables, averaging 1.04 and 2.6 columns per table respectively. Map\&Make, in contrast, misses out on 0 and 0.21 columns in the Team and Player tables respectively. Map\&Make also produces extra columns for every table capturing the non-statistical aspects of entities, which are absent in the Ground Truths. 


\begin{figure}[h]
    \centering
    \includegraphics[width=0.48\textwidth]{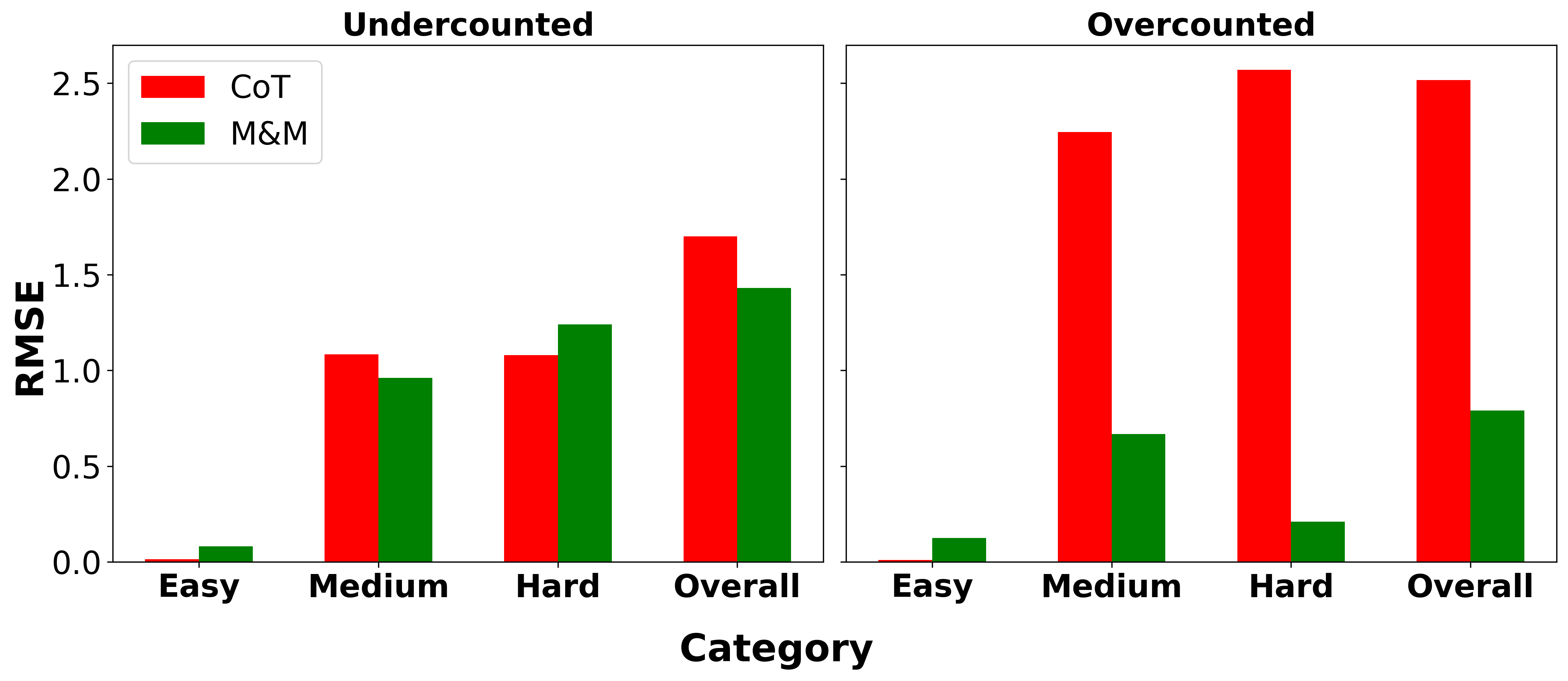}
    \caption{\small RMSE of Overcounting and Undercounting Instances for Livesum. Uncercounted refers to cell values less than the ground truth, Overcounted refers to cell values more than the ground truths.}
    \label{fig:livesum-trend}
\end{figure}
\textbf{Livesum:} 
We further segregate instances of overcounting and undercounting of events separately to analyze where LLMs miss out on information (undercounting) and hallucinate (overcount). Figure \ref{fig:livesum-trend} shows the segregated RMSE scores across both instances. We observe over 67\% of the total errors for CoT are hallucinations. These findings align with other research the demonstrate LLM's struggle with counting tasks.\cite{ball2024can};\cite{zhang2024counting}. Contrastively, our methodology shows stable performance across all difficulty levels with overcounting RMSE decreasing by a big margin.

\subsection{Analysis on Efficiency and Table Planning}
\begin{figure}[h]
    \centering
    \includegraphics[width=0.48\textwidth]{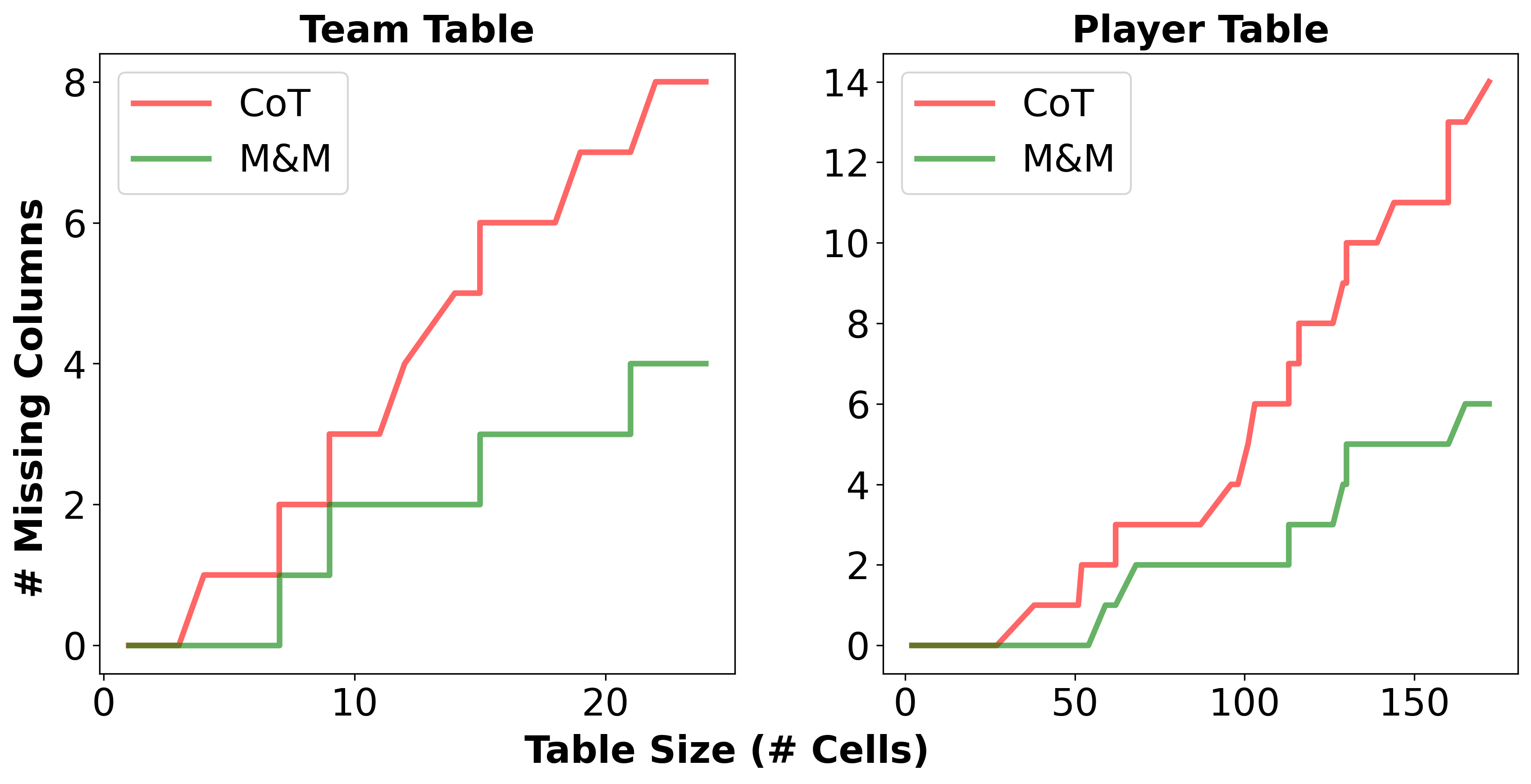}
    \caption{\small Comparison of Schema-Coverage with Increasing Table Sizes for Rotowire.}
    \label{fig:rotowire-trend}
    \vspace{-1.0em}
\end{figure}
Figure \ref{fig:rotowire-trend} describes the schemas coverage of CoT and our method's extracted schemas. We see a very high correlation between missing columns and Ground Truth tables for CoT, showcasing poor performance as schema size increases. We show stable performance across larger table sizes for both player and team tables.

The facets of information coverage and accurate extraction in structured summarization tasks are challenging due to the high variability of linguistic structures, implicit relationships, and contextual dependencies present in unstructured text. These complexities, including syntactic ambiguities, coreference resolution challenges, and implicit entity attributions, make it difficult for models to reliably infer and represent structured information while preserving the fidelity of the original content.
Rotowire tests the capabilities of models in planning exhaustive Table schemas from highly contexualised narratives where as Livesum tests exacting occurrences of different events. 
M\&M prompting shows significant improvements across both datasets and hence, is a more reliable and exhaustive table generation strategy.
\section{Ablation Study}

\begin{table}[h!]
\centering
\resizebox{\linewidth}{!}{%
\begin{tabular}{lcccccc}
\hline \hline
\multicolumn{1}{c}{\multirow{2}{*}{\textbf{M\&M}}} &
  \multicolumn{2}{c}{\textbf{Correctness}} &
  \multicolumn{2}{c}{\textbf{Completeness}} &
  \multicolumn{2}{c}{\textbf{Overall}} \\ \cline{2-7} 
\multicolumn{1}{c}{} &
  \textbf{Team} &
  \textbf{Player} &
  \textbf{Team} &
  \textbf{Player} &
  \textbf{Team} &
  \textbf{Player} \\ \hline
\multicolumn{7}{c}{\textbf{Unified}} \\ \hline
\textbf{No Ablation} &
  \textbf{75.27} &
  \textbf{91.17} &
  \textbf{77.47} &
  \textbf{92.92} &
  \textbf{74.80} &
  \textbf{91.36} \\
\textbf{- Atomization} &
  72.91 &
  82.80 &
  77.37 &
  85.61 &
  73.28 &
  82.42 \\
\textbf{- Iterative} &
   &
   &
   &
   &
   &
   \\
\textbf{\hspace{2em}Schema} &
  69.74 &
  80.45 &
  70.75 &
  86.42 &
  68.50 &
  81.73 \\
\textbf{\hspace{2em}Table} &
  70.58 &
  83.38 &
  69.18 &
  89.05 &
  67.55 &
  84.79 \\ \hline
\multicolumn{7}{c}{\textbf{3-Step}} \\ \hline
\textbf{No Ablation} &
  65.89 &
  83.54 &
  79.49 &
  92.43 &
  69.95 &
  \textbf{86.71} \\
\textbf{- Atomization} &
  \textbf{73.43} &
  \textbf{84.39} &
  77.62 &
  88.52 &
  \textbf{73.41} &
  84.87 \\
\textbf{- Iterative} &
   &
   &
   &
   &
   &
   \\
\textbf{\hspace{2em}Schema} &
  69.48 &
  80.72 &
  \textbf{67.99} &
  \textbf{90.63} &
  65.85 &
  83.91 \\
\textbf{\hspace{2em}Table} &
  63.98 &
  77.39 &
  \textbf{80.26} &
  \textbf{93.69} &
  68.84 &
  83.12 \\ \hline \hline
\end{tabular}%
}
\caption{Ablations on Rotowire.}
\label{tab:ablation-rotowire}
\vspace{-1.0em}
\end{table}

To understand different components our method, we conducted some additional studies, where we focused on removing a key step from our method, showcased in Table~\ref{tab:ablation-rotowire} \& \ref{tab:livesum-ablation}.
\subsection{Rotowire}
We observed that removing Atomization step leads to a drop in the Completeness of both the Team and the Player table across both Unfied and 3-Step variation. This demonstrates that breaking down multi-entity complex statements is essential for information coverage. Ablating the Iterative Schema Generation step consistently degrades the performance, which illustrates that it is a crucial step in our method, as this is effective for tabular structure and enables better table filling as manifested. Furthermore, removing Iterative Table Generation step, we observe a significant drop in the Correctness and Overall metrics, however it is to be noted that there is slight increase in the Completeness for 3-Step. This suggests that without iterative updates there is less information captured and introduces redundancy. 

\subsection{Livesum}
\label{sec:livesum-ablation}
\begin{table}[h!]
\centering
\resizebox{\linewidth}{!}{%
\begin{tabular}{lcccccccc}
\hline \hline
\multicolumn{1}{c}{\multirow{2}{*}{\textbf{M\&M}}} &
  \multicolumn{2}{c}{\textbf{Easy}} &
  \multicolumn{2}{c}{\textbf{Medium}} &
  \multicolumn{2}{c}{\textbf{Hard}} &
  \multicolumn{2}{c}{\textbf{Average}} \\ \cline{2-9} 
\multicolumn{1}{c}{} &
  \textbf{RMSE} & \textbf{ER} &
  \textbf{RMSE} & \textbf{ER} &
  \textbf{RMSE} & \textbf{ER} &
  \textbf{RMSE} & \textbf{ER} \\ \hline
\multicolumn{9}{c}{\textbf{Unified}} \\ \hline
\textbf{No Ablation} & 0.08 & 3.84 & 0.87 & 31.34 & 2.63 & 80.29 & 1.47 & 35.00 \\
\textbf{- Atomization} 
   & 0.05 & 2.73
   & 0.53 & 18.58
   & 1.15 & 44.37
   & 0.75 & 21.06 \\
\textbf{- Iterative} &
  0.02 & 0.93 & 1.52 & 54.33 & 3.41 & 86.20 & 2.08 & 48.94  \\ \hline
\multicolumn{9}{c}{\textbf{3-Step}} \\ \hline
\textbf{No Ablation} &
  0.1 & 4.54 & 0.6 & 23.74 & 0.89 & 33.14 & 0.70 & 21.37 \\
\textbf{- Atomization} &
  0.05 & 2.90 & 
  0.64 & 20.36 & 
  1.59 & 47.89 &  
  1.01 & 22.96   \\
\textbf{- Iterative} &
  0.08 & 3.73 & 1.47 & 37.44 &
  2.81 & 65.67 & 1.85 & 36.07 \\ \hline \hline
\end{tabular}%
}
\caption{Ablations on Livesum}
\label{tab:livesum-ablation}
\end{table}
We perform the following ablations of our 3-step (M\&M - 3S) and unified (M\&M - U) approach. Since the schema for every table is the same for every test sample, we don't ablate the iterative structure step for this study. We see a significant decrease in performance without the iterative table-filling step across both Unified and 3-step settings. This underscores the importance of iterative table filling as models can more accurately track dynamic updates without hallucinating or missing out on events. Interestingly, we see a small performance increase after removing the unified variant's atomization step. This is because in a one-shot setting, removing the atomization part boils the problem down to a fixed schema updation.

\section{On Broader Domain Generalization \& Multi-Table Generation - A study on \textbf{Wiki40B}}
As the livesum and rotowire benchmarks both belong to the sports domain, we conduct a study to validate the effectiveness of M\&M to open domain texts. Taking inspiration from the pioneering work of \cite{jain-etal-2024-structsum}, we extend our evaluation to Wiki40B, a large-scale, open-domain, multilingual (40 languages) corpus of  Wikipedia articles from several domains. 
Wiki40B poses a significant challenge for text-to-table generation due to its wide topical range and the unstructured nature of its content. Articles span a wide variety of topics, from historical revolutions to astrophysics, each with a distinct narrative style. 
Specifically:
\begin{itemize}
    \item We sample \textbf{500} English articles, each containing at least \textbf{30} numerical values and \textbf{3+} sentences, to ensure sufficient complexity.
    \item Due to the absence of gold-standard tables, we adopt \textbf{AutoQA}, a reference-less evaluation metric that tests whether the generated table can accurately answer factoid questions derived from the original article.
\end{itemize}
In the table below, we report AutoQA scores on Wiki40B for different methods.
\begin{table}[h!]
\centering
\small
\begin{tabular}{l c}
\toprule
\textbf{Method} & \textbf{AutoQA-Score} \\
\midrule
Chain of Thoughts & 63.68 \\
Text-Tuple-Table & 67.07 \\
Map \& Make (3-step) & \textbf{76.40} \\
\bottomrule
\end{tabular}
\caption{AutoQA scores using GPT-4o in zero-shot settings.}
\label{tab:autoqa-results}
\end{table}

Map\&Make achieves a \textbf{14\% improvement over CoT} and a \textbf{9\% gain over T\textsuperscript{3}}, demonstrating its \textbf{generalization capacity to open-domain, non-synthetic text}. Moreover, as compared to generating a fixed number of tables, we observe our framework dynamically adapts to multiple table generation (Range \textbf{1-13}; Mean: \textbf{6.02}; Std: \textbf{4.33}) to improve comprehension.
These results suggest that our modular framework, especially the atomization and iterative schema construction, scales effectively to structurally diverse open-domain text, making it a generalised and interpreted framework.

\section{Related Works}
With the advent of Large Language Models (LLMs), numerous methods for information extraction and summarization have been developed. For extraction tasks that involve structured data, two fields have been researched; Table-to-Text and Text-to-Table have been researched, with the latter being less explored. 

\textbf{Table-to-Text} Prior works have been referred to as data-to-text in a more general sense. Former works in table-to-text traditionally uses sequence-to-sequence (seq2seq) models to generate table descriptions \cite{lebret-etal-2016-neural}, \cite{wiseman-etal-2017-challenges}, \cite{wikibio}, \cite{wang-etal-2020-towards}. Since there exists a bidirectional relationship between structural data and unstructured text, the inverse task text-to-table has also started gained attention.

\textbf{Text-to-Table} Recent studies investigates diverse methods to transform unstructured text into structured tables while addressing the underlying issues of schema inference, handling complex data relations, and knowledge integration. A key contribution to this area has been covered by \cite{text-2-table-2022-acl}, who introduced the approach of information extraction. They developed a seq2seq model with a fine-tuned version of BART for text-to-table translation using the Rotowire dataset. Additionally, studies like \cite{li2023sequence}; \cite{sundar2024gtbls} have explored more sophisticated methods recognising challenges of interpreting 2-d structures as a linearized sequence object. While these methods outperforms traditional methods that employ relation extraction~\cite{zheng-etal-2017-joint}, \cite{zeng-etal-2018-extracting}, \cite{luan-etal-2019-general}, \cite{zhong-chen-2021-frustratingly} and named entity recognition (NER)~\cite{huang2015bidirectional}, \cite{ma-hovy-2016-end}, \cite{lample-etal-2016-neural}, \cite{devlin-etal-2019-bert}, there has been less exploration of this task on leveraging the emergent capabilities of LLMs in developing a generalisable methodology. Our work builds upon these foundations while addressing key limitations. We introduce a schema-agnostic approach that can dynamically infer table structures from open-domain text. Our method aims to provide a more flexible and comprehensive solution to the text-to-table generation task.
\vspace{-0.2em}
\section{Conclusion}
\vspace{-0.5em}
We introduce Map\&Make, a versatile approach to extract complex information from unstructured text. M\&M tackles the challenges by breaking down the text into its atomic statements, this decomposition then helps extract the latent schema, which ultimately aids in populating the tables. M\&M's core strength lies in its ability to dynamically infer schema from open-domain language, as well as its ability to handle complex information such as numerical aggregation and detailed qualitative descriptions. Furthermore, our corrections of the Rotowire benchmark provides a more reliable and fairer evaluation platform for future research in this domain. Our meticulous evaluation on Rotowire, Livesum, and Wiki40B showcases improvements in information coverage and localization. M\&M mitigates information loss and maintains stable performance even with a larger input context length.

\vspace{1em}

\section*{Limitations}
While Map\&Make demonstrates significant advancements in schema-guided text-to-table generation, it is important to acknowledge certain limitations.

\noindent\textbf{Reliance on LLMs} M\&M's performance is tied to the text generation capabilities of the LLMs, the quality of atomization, schema extraction, and table generation/filling is directly proportional to the LLM's reasoning abilities. This leads to several dependencies such as handling highly ambiguous text, grammatical errors, and contradictory information.

\noindent\textbf{Computational Cost} While the iterative nature helps in information coverage as seen in Table~\ref{tab:ablation-rotowire} and Table~\ref{tab:rotowire-llm-based}, but becomes computationally expensive due to lengthy outputs, such as those in Livesum. We explore fine-tuning smaller models like Llama 8B to reduce computational overhead, and we find that, while it improves performance, there is a significant gap between smaller fine-tuned models and larger SoTA LLMs like GPT-40, Gemini-2.0-flash. More details about results and experiments can be found in appendix \ref{sec:appendix-finetune}.

\noindent\textbf{Evaluation metric limitation} While we cover a wide range of metrics, these metrics may not perfectly capture all nuanced aspects of the information in the table, thus a subjective human evaluation is needed to test the practicality of our method.

\noindent\textbf{Hierarchial Structures} This methodology does not apply to tabular structures with free-form structural complexities such as hierarchical headers, cells with merged cells. \cite{cheng2021hitab}. These limitations highlight areas for future research and development to further enhance the robustness, efficiency, and generalizability of schema-guided text-to-table generation frameworks like Map\&Make.


\section*{Ethics Statement}
The authors ensure that this work meets the highest ethical standards in research and publication. We have carefully addressed ethical considerations to guarantee appropriate behavior and fair use of computational linguistics methods. Our assertions are consistent with experimental data. While some stochasticity is predicted with black-box Large Language Models, we minimize it by keeping a constant temperature, top\_p, top\_k. Furthermore, the use of LLMs such as GPT-4o, Gemini, and Llama in this study adheres to their policies of usage. We have used AI assistants (Grammarly and ChatGPT) to address the grammatical errors and rephrase the sentences. Finally, to the best of our knowledge, we believe that this work introduces no additional risk.

\section*{Acknowledgments}
We thank the Complex Data Analysis and Reasoning Lab at Arizona State University for computational support.We thank the anonymous reviewers for their helpful feedback.


\clearpage

\appendix
\section*{\centering Appendices}
\addcontentsline{toc}{section}{Appendices}
\renewcommand{\thesection}{\Alph{section}}

\section{Datasets}
In this section, we present more details about the datasets used in this study.
\subsection{Livesum}
\label{subsec:appendix}
\begin{figure}[h]
    \centering
    \includegraphics[width=0.48\textwidth]{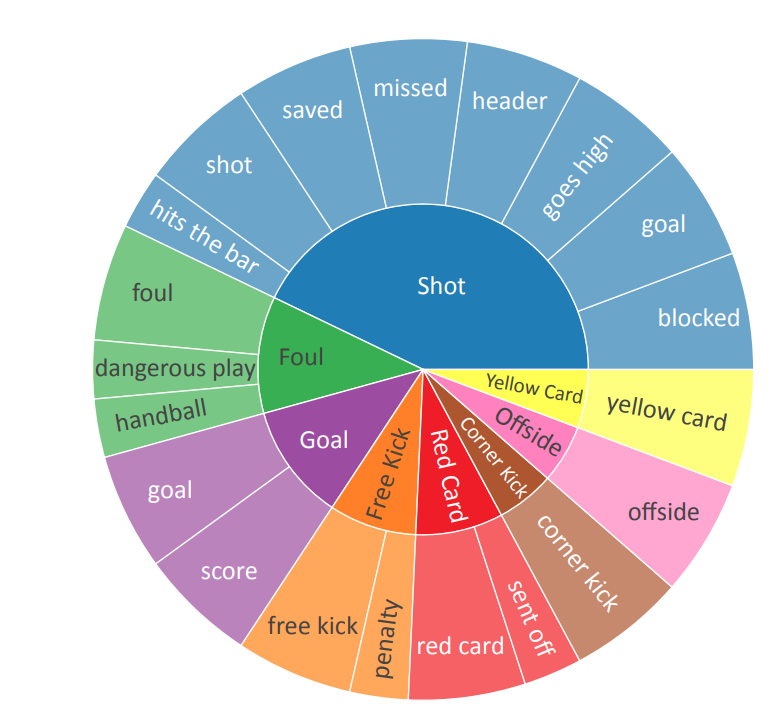}
    \caption{\small Eight types of event information (inner circle)
that require summarization in Livesum dataset, along
with their common expressions (outer circle) in the commentary. \cite{t3}}
    \label{fig:appendix-image}
\end{figure}

This benchmark was created by scraping live commentory of football games played in the English Premier League. Player Names and Team Names were anonymised and summary tables were annotated manually. The difficulty level of each event (columns in a table) is determined by the number of different descriptions that are present for each event, as shown in figure \ref{fig:appendix-image}. Goals and Red Cards are classified as easy due to their rare occurrences and straightforward descriptions. Shots and Fouls are classified as Hard as they have the most varying descriptions. The remaining four events are classified as Medium difficulty.

\label{sec:Livesum Characteristics}

\begin{figure*}[ht]
\centering
\includegraphics[width=0.95\linewidth]{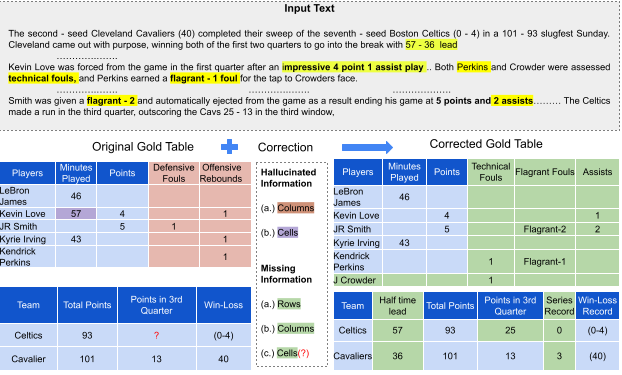}
\caption{Instances of hallucinations and missing information in Rotowire.}
\label{fig:corrections}
\end{figure*}

\subsection{Rotowire}
The Rotowire dataset was originally introduced by \cite{wiseman-etal-2017-challenges} for table-to-text generation, comprising NBA game summaries paired with detailed box and line scores. It was later repurposed by \cite{text-2-table-2022-acl} for text-to-table generation by filtering the original tables to retain only records that could be grounded in the textual summaries. For each sample summary, a team-level and a player-level table were constructed. The benchmark consists of 3,398 training, 727 validation, and 728 test samples. During our experiments, we observed multiple table entries with no grounding in the corresponding text, which we corrected to have a clean and reliable benchmark.
\subsubsection{Dataset Corrections}
\label{sec:rotowire-corrections}
The match summaries often contain information about player and team performances outside the scope of the match in question. For example, the recent performance trends of teams, upcoming fixtures, players on impressive streaks, etc.   
To distill relevant information, we operate the instruction to keep only \textit{statistical values of Teams and Players pertinent to the match in question.} \\ 
First, we extract the global column headers from the complete test set for the Team and Player tables. We further add two column headers to the global set "Points in the Paint" and "Half-Time Score". The former represents the points made by the team through 2-point shots, and the latter represents the score of both teams at half-time, hence relevant to the team performance summary. Row Headers require no correction at the global level as they contain Player Names and Team Names, hence are validated for every sample individually.  

Every table follows the same structure with Row Headers as the teams (Home Team, Away Team) and 8 columns, one for each event. As our problem statement involves generating summary tables without any schematic constraints, we find extra events such as \textbf{Passes, Assists, Through Balls} that are contextually relevant but cannot be evaluated using the ground truth. Hence, we employ a paraphrasing agent to transform our outputs as per the ground truth's structure. Full prompt used for paraphrasing can be found in appendix \ref{sec:rotowire-atomic}

\section{Additional Results}
\subsection{Rotowire}
\label{sec:Why precision decreases}
\begin{table}[ht!]
  \centering
  \resizebox{\linewidth}{!}{%
  \begin{tabular}{lccccccccc}
  \hline \hline
  \multicolumn{1}{c}{\multirow{2}{*}{\textbf{Method}}} &
    \multicolumn{3}{c}{\textit{\textbf{EM}}} &
    \multicolumn{3}{c}{\textit{\textbf{CHRF}}} &
    \multicolumn{3}{c}{\textit{\textbf{BERT}}} \\ \cline{2-10} 
  \multicolumn{1}{c}{} &
    \multicolumn{1}{c}{\textbf{Cell}} &
    \multicolumn{1}{c}{\textbf{Row}} &
    \multicolumn{1}{c}{\textbf{Col}} &
    \multicolumn{1}{c}{\textbf{Cell}} &
    \multicolumn{1}{c}{\textbf{Row}} &
    \multicolumn{1}{c}{\textbf{Col}} &
    \multicolumn{1}{c}{\textbf{Cell}} &
    \multicolumn{1}{c}{\textbf{Row}} &
    \multicolumn{1}{c}{\textbf{Col}} \\ \hline
  \multicolumn{1}{c}{\textit{\textbf{}}} & \multicolumn{9}{c}{\textit{\textbf{GPT-4o Zero-Shot}}}                                                                      \\ \hline
  \textbf{CoT}                           & 18.52 & 47.28 & 27.89 & 34.10 & 77.60 & 46.06 & 34.92 & 57.19 & 66.58 \\
  \textbf{M\&M - U} &
    \multicolumn{1}{c}{15.07} &
    \multicolumn{1}{c}{46.43} &
    \multicolumn{1}{c}{21.73} &
    \multicolumn{1}{c}{29.92} &
    \multicolumn{1}{c}{74.59} &
    \multicolumn{1}{c}{41.48} &
    \multicolumn{1}{c}{34.34} &
    \multicolumn{1}{c}{57.54} &
    \multicolumn{1}{c}{59.56} \\ \hline
  \multicolumn{1}{c}{\textit{\textbf{}}} & \multicolumn{9}{c}{\textit{\textbf{GPT-4o One-Shot}}}                                                                       \\ \hline
  \textbf{CoT}                           & 56.62 & 88.44 & 61.20 & 76.16 & 92.36 & 79.50 & 83.01 & 91.41 & 87.20 \\
  \textbf{M\&M - U}                      & 45.41 & 88.39 & 50.38 & 61.24 & 91.38 & 68.77 & 68.52 & 93.37 & 74.50 \\
  \textbf{M\&M - 3S}                     & 23.30 & 58.33 & 34.69 & 40.17 & 79.36 & 53.71 & 43.05 & 67.41 & 60.87 \\ \hline
  \multicolumn{1}{l}{}                   & \multicolumn{9}{c}{\textit{\textbf{Gemini-2.0 Zero-Shot}}}                                                                  \\ \hline
  \textbf{CoT}                           & 33.56 & 21.90 & 51.85 & 37.71 & 76.49  & 49.09 & 40.46 & 58.64 & 79.93 \\
  \textbf{M\&M - U} &
    \multicolumn{1}{l}{16.85} &
    \multicolumn{1}{l}{45.41} &
    \multicolumn{1}{l}{28.03} &
    \multicolumn{1}{l}{36.39} &
    \multicolumn{1}{l}{71.97} &
    \multicolumn{1}{l}{48.98} &
    \multicolumn{1}{l}{36.91} &
    \multicolumn{1}{l}{56.66} &
    \multicolumn{1}{l}{61.78} \\ \hline
  \multicolumn{1}{c}{\textit{\textbf{}}} & \multicolumn{9}{c}{\textit{\textbf{Gemini-2.0 One-Shot}}}                                                                   \\ \hline
  \textbf{CoT}                           & 56.27 & 91.79 & 59.85 & 75.56 & 94.13 & 79.10 & 84.18 & 95.95 & 85.90 \\
  \textbf{M\&M - U}                      & 57.00 & 90.32 & 61.61 & 71.77 & 93.20 & 77.64 & 77.90 & 94.76 & 80.38 \\
  \textbf{M\&M - 3S}                     & 49.90 & 86.11 & 55.95 & 62.49 & 89.82 & 71.86 & 69.86 & 91.52 & 74.65 \\ \hline \hline
  \end{tabular}%
  }
  \caption{Rotowire Correctness Scores}
  \label{tab:appendix precision}
  \end{table}

\begin{table}[h!]
\centering
\resizebox{\linewidth}{!}{%
\begin{tabular}{lccccccccc}
\hline
\multicolumn{1}{c}{\multirow{2}{*}{\textbf{Method}}} &
  \multicolumn{3}{c}{\textit{\textbf{EM}}} &
  \multicolumn{3}{c}{\textit{\textbf{CHRF}}} &
  \multicolumn{3}{c}{\textit{\textbf{BERT}}} \\ \cline{2-10} 
\multicolumn{1}{c}{} &
  \textbf{Cell} &
  \textbf{Row} &
  \textbf{Col} &
  \textbf{Cell} &
  \textbf{Row} &
  \textbf{Col} &
  \textbf{Cell} &
  \textbf{Row} &
  \textbf{Col} \\ \hline
\multicolumn{1}{c}{\textit{\textbf{}}} &
  \multicolumn{9}{c}{\textit{\textbf{GPT-4o Zero-Shot}}} \\ \hline
\textbf{CoT} &
  18.93 &
  48.71 &
  28.36 &
  35.56 &
  78.90 &
  46.87 &
  35.49 &
  58.92 &
  61.74 \\
\textbf{M\&M - U} &
  16.61 &
  47.79 &
  24.69 &
  33.72 &
  76.16 &
  45.87 &
  37.20 &
  59.25 &
  61.29 \\ \hline
\multicolumn{1}{c}{\textit{\textbf{}}} &
  \multicolumn{9}{c}{\textit{\textbf{GPT-4o One-Shot}}} \\ \hline
\textbf{CoT} &
  38.93 &
  87.99 &
  42.22 &
  53.22 &
  91.95 &
  58.93 &
  65.69 &
  91.09 &
  69.47 \\
\textbf{M\&M - U} &
  46.85 &
  89.05 &
  51.71 &
  63.08 &
  92.01 &
  70.21 &
  70.33 &
  93.92 &
  75.47 \\
\textbf{M\&M - 3S} &
  26.66 &
  60.28 &
  40.20 &
  46.60 &
  81.37 &
  60.41 &
  48.05 &
  69.48 &
  66.81 \\ \hline
 &
  \multicolumn{9}{c}{\textit{\textbf{Gemini-2.0 Zero-Shot}}} \\ \hline
\textbf{CoT} &
  20.72 &
  49.02 &
  30.30 &
  35.68 &
  77.91 &
  45.55 &
  38.33 &
  60.34 &
  64.07 \\
\textbf{M\&M - U} &
  17.71 &
  46.09 &
  28.91 &
  38.00 &
  72.82 &
  50.24 &
  37.97 &
  57.81 &
  60.42 \\ \hline
\multicolumn{1}{c}{\textit{\textbf{}}} &
  \multicolumn{9}{c}{\textit{\textbf{Gemini-2.0 One-Shot}}} \\ \hline
\textbf{CoT} &
  \multicolumn{1}{r}{45.65} &
  \multicolumn{1}{r}{91.36} &
  \multicolumn{1}{r}{48.71} &
  \multicolumn{1}{r}{61.78} &
  \multicolumn{1}{r}{93.75} &
  \multicolumn{1}{r}{66.89} &
  \multicolumn{1}{r}{71.66} &
  \multicolumn{1}{r}{95.59} &
  \multicolumn{1}{r}{73.46} \\
\textbf{M\&M - U} &
  \multicolumn{1}{r}{56.03} &
  \multicolumn{1}{r}{90.24} &
  \multicolumn{1}{r}{60.63} &
  \multicolumn{1}{r}{70.75} &
  \multicolumn{1}{r}{93.08} &
  \multicolumn{1}{r}{76.56} &
  \multicolumn{1}{r}{76.44} &
  \multicolumn{1}{r}{94.66} &
  \multicolumn{1}{r}{78.89} \\
\textbf{M\&M - 3S} &
  \multicolumn{1}{r}{53.63} &
  \multicolumn{1}{r}{87.14} &
  \multicolumn{1}{r}{59.89} &
  \multicolumn{1}{r}{66.88} &
  \multicolumn{1}{r}{90.82} &
  \multicolumn{1}{r}{75.61} &
  \multicolumn{1}{r}{73.32} &
  \multicolumn{1}{r}{92.30} &
  \multicolumn{1}{r}{77.74} \\ \hline \hline
\end{tabular}%
}
\caption{Rotowire Overall Scores}
\label{tab:appendix f1}
\end{table}
In Table \ref{tab:appendix precision} and \ref{tab:appendix f1}, we provide the Correctness (Precision) and Overall Scores (F1) for Rotowire on GPT-4o, and Gemini-2.0-flash-exp. We see minimal to no improvement in correctness scores across similarity metrics due to the lack of any statistical attributes in the ground-truths.  Map \& Make, being an unsupervised designed to enhance coverage of information produces descriptive columns such as \textbf{Injury Status}, categorical columns such as \textbf{Starting 5 player/ Off the Bench Player} (used to describe whether a player is present in the game from the start of the match or is substituted mid-game). Also, columns that represent attributes not directly relevant to the game such as \textbf{Average Points Scored past 3 games}, \textbf{Win Streak} have no reference in ground truths which leads to penalisation of our methodology. However, to validate the correctness of the extra columns generated in our tables we employ AutoQA (shown in Table \ref{tab:rotowire-llm-based}). M\&M's consistent improvement on this metric validates the faithfulness of extra information to the gold text.

\section{Fine-Tuning Study on Smaller LLMs}
\label{sec:appendix-finetune}

To evaluate whether our modular Map\&Make framework can be used to train smaller language models with improved reasoning and structured generation capabilities, we conducted a fine-tuning study. We used our pipeline on the \textsc{Rotowire} training corpora to generate high-quality supervision data using \textbf{Gemini 2.0}. These structured table-text pairs were then used to fine-tune a \textbf{LLaMA 3 8B Instruct} model.

We evaluated the fine-tuned model on the \textsc{Rotowire} test set using the \textsc{TabEval} and \textsc{AutoQA} metrics. We compare its performance with a Chain-of-Thought (CoT) baseline and our M\&M (3-step) framework. Results are shown below.

\vspace{0.5em}
\begin{table}[h]
\centering
\small
\setlength{\tabcolsep}{4.5pt} 
\begin{tabular}{lcccccc}
\toprule
\textbf{Method} & \multicolumn{2}{c}{\textbf{Precision}} & \multicolumn{2}{c}{\textbf{Recall}} & \multicolumn{2}{c}{\textbf{F1}} \\
\cmidrule(lr){2-3} \cmidrule(lr){4-5} \cmidrule(lr){6-7}
 & Team & Player & Team & Player & Team & Player \\
\midrule
Fine-tune   & 60.16 & 59.84 & 41.61 & 61.58 & 45.62 & 58.23 \\
CoT         & 80.23 & 86.74 & 56.26 & 57.74 & 64.34 & 65.57 \\
M\&M-3S     & 48.78 & 74.22 & 61.43 & \textbf{87.34} & 51.01 & \textbf{78.56} \\
\bottomrule
\end{tabular}
\caption{TabEval scores on \textsc{Rotowire}. Player columns reflect per-entity evaluation; Team reflects aggregated rows.}
\label{tab:appendix-tabeval}
\end{table}

\begin{table}[h]
\centering
\small
\setlength{\tabcolsep}{6pt}
\begin{tabular}{lc}
\toprule
\textbf{Method} & \textbf{AutoQA Accuracy} \\
\midrule
Fine-tune   & 48.65 \\
CoT         & 41.42 \\
M\&M-3S     & \textbf{80.38} \\
\bottomrule
\end{tabular}
\caption{AutoQA scores using QA pairs generated by Gemini 2.0.}
\label{tab:appendix-autoqa}
\end{table}
\vspace{-0.4em}
These results show that fine-tuning smaller LLMs with data generated by Gemini 2.0 can modestly improve performance on structured prediction. However, limitations remain:

\begin{itemize}
\vspace{-0.7em}
\item \textbf{Instruction alignment:} Smaller LLMs struggle with consistent table formatting and multi-row outputs.
\vspace{-0.7em}
\item \textbf{Context limitations:} Shorter context lengths hinder long-document comprehension and global reasoning.
\vspace{-0.7em}
\item \textbf{Training quality:} Datasets like Rotowire contain hallucinations. Gemini 2.0 outputs offered improved fidelity during fine-tuning.
\vspace{-1em}
\end{itemize}

This study illustrates that M\&M can bootstrap fine-tuning for smaller models and serve as a scalable foundation for schema-agnostic table generation in low-resource or multilingual settings.

\onecolumn
\section{M\&M Sample Outputs}
\subsection{Rotowire}
\subsubsection{Propositional Atomization}
\begin{tcolorbox}[colback=black!10, colframe=black!80, title=M\&M Rotowire Model Output - Atomic, breakable]
\small
\begin{Verbatim}[breaklines=true, fontsize=\scriptsize]
***REASONING STEPS***:
For each sentence, identify the entities and their corresponding events.

**Sentence 1**: The Hornets (5 - 15) snapped a 10 - game winning streak and sent the Knicks (4 - 17) to their seventh straight loss with Friday’s 103 - 102 victory.
- Entities: The Hornets, The Knicks
- Events: Team records, game result, winning streak, losing streak
*Atomic sentences*:
The Hornets' record is 5 wins.
The Hornets' record is 15 losses.
The Hornets snapped a 10-game winning streak.
The Knicks' record is 4 wins.
The Knicks' record is 17 losses.
The Knicks have a seven-game losing streak.
The Hornets defeated the Knicks 103-102 on Friday.

**Sentence 2**: Charlotte, which led by 14 entering the fourth quarter, allowed the Knicks to creep back into the game in the final four minutes.
- Entities: Charlotte, The Knicks
- Events: Lead, game progression
*Atomic sentences*:
Charlotte led by 14 points entering the fourth quarter.
The Knicks reduced Charlotte's lead in the final four minutes.

**Sentence 3**: Back-to-back threes from J.R. Smith and Carmelo Anthony brought New York within two, and another Anthony three less than a minute later gave the Knicks a one - point advantage with 40 seconds to play.
- Entities: J.R. Smith, Carmelo Anthony, The Knicks
- Events: Three-point shots, score change
*Atomic sentences*:
J.R. Smith made a three-point shot.
Carmelo Anthony made a three-point shot.
The Knicks were within two points of Charlotte.
Carmelo Anthony made another three-point shot.
The Knicks had a one-point advantage with 40 seconds to play.

**Sentence 4**: Following missed baskets on both ends, the Hornets had the ball with four seconds to play with a chance to win.
- Entities: The Hornets
- Events: Missed baskets, possession
*Atomic sentences*:
Both teams missed baskets.
The Hornets had possession of the ball with four seconds to play.

**Sentence 5**: The ball was inbounded to Kemba Walker, who drove into the paint and laid it in at the buzzer for his 11th point of the game.
- Entities: Kemba Walker
- Events: Inbound pass, scoring
*Atomic sentences*:
The ball was inbounded to Kemba Walker.
Kemba Walker drove into the paint.
Kemba Walker scored at the buzzer.
Kemba Walker scored his 11th point of the game.

**Sentence 6**: The Hornets were led by 22 points from Gerald Henderson and 16 points and 13 boards from Al Jefferson.
- Entities: Gerald Henderson, Al Jefferson
- Events: Scoring, rebounds
*Atomic sentences*:
Gerald Henderson scored 22 points.
Al Jefferson scored 16 points.
Al Jefferson had 13 rebounds.

**Sentence 7**: Lance Stephenson added 16 points (7 - 15 FG) to go with six assists and six rebounds, and Gary Neal chipped in 13 points off the bench.
- Entities: Lance Stephenson, Gary Neal
- Events: Scoring, assists, rebounds
*Atomic sentences*:
Lance Stephenson scored 16 points.
Lance Stephenson made 7 of 15 field goal attempts.
Lance Stephenson had six assists.
Lance Stephenson had six rebounds.
Gary Neal scored 13 points off the bench.

**Sentence 8**: The Hornets shot 48 percent from the floor but made just 22 percent of their three - point attempts (4 - 18 3Pt).
- Entities: The Hornets
- Events: Shooting percentages
*Atomic sentences*:
The Hornets shot 48 percent from the floor.
The Hornets made 22 percent of their three-point attempts.
The Hornets made 4 of 18 three-point attempts.

**Sentence 9**: New York got 32 points from Anthony to go with seven rebounds and four assists.
- Entities: Carmelo Anthony
- Events: Scoring, rebounds, assists
*Atomic sentences*:
Carmelo Anthony scored 32 points.
Carmelo Anthony had seven rebounds.
Carmelo Anthony had four assists.

**Sentence 10**: The only other Knicks to score in double figures were Tim Hardaway, Jr. and J.R. Smith, who had 13 and 14 points, respectively.
- Entities: Tim Hardaway, Jr., J.R. Smith
- Events: Scoring
*Atomic sentences*:
Tim Hardaway, Jr. scored 13 points.
J.R. Smith scored 14 points.

**Sentence 11**: Smith also added five assists and four boards.
- Entities: J.R. Smith
- Events: Assists, rebounds
*Atomic sentences*:
J.R. Smith had five assists.
J.R. Smith had four rebounds.

**Sentence 12**: The Knicks were out - rebounded 41 - 34 but committed only six turnovers, three of which were charged to Samuel Dalembert.
- Entities: The Knicks, Samuel Dalembert
- Events: Rebounds, turnovers
*Atomic sentences*:
The Knicks were out-rebounded 41 to 34.
The Knicks committed six turnovers.
Samuel Dalembert committed three turnovers.

**Sentence 13**: Up next, the Knicks host the Blazers on Sunday.
- Entities: The Knicks, The Blazers
- Events: Upcoming game
*Atomic sentences*:
The Knicks will host the Blazers on Sunday.

**Sentence 14**: Charlotte is off until Wednesday, when they face the Celtics.
- Entities: Charlotte, The Celtics
- Events: Upcoming game
*Atomic sentences*:
Charlotte is off until Wednesday.
Charlotte will face the Celtics on Wednesday.

### Atomic Statements:
The Hornets' record is 5 wins.
The Hornets' record is 15 losses.
The Hornets snapped a 10-game winning streak.
The Knicks' record is 4 wins.
The Knicks' record is 17 losses.
The Knicks have a seven-game losing streak.
The Hornets defeated the Knicks 103-102 on Friday.
Charlotte led by 14 points entering the fourth quarter.
The Knicks reduced Charlotte's lead in the final four minutes.
J.R. Smith made a three-point shot.
Carmelo Anthony made a three-point shot.
The Knicks were within two points of Charlotte.
Carmelo Anthony made another three-point shot.
The Knicks had a one-point advantage with 40 seconds to play.
Both teams missed baskets.
The Hornets had possession of the ball with four seconds to play.
The ball was inbounded to Kemba Walker.
Kemba Walker drove into the paint.
Kemba Walker scored at the buzzer.
Kemba Walker scored his 11th point of the game.
Gerald Henderson scored 22 points.
Al Jefferson scored 16 points.
Al Jefferson had 13 rebounds.
Lance Stephenson scored 16 points.
Lance Stephenson made 7 of 15 field goal attempts.
Lance Stephenson had six assists.
Lance Stephenson had six rebounds.
Gary Neal scored 13 points off the bench.
The Hornets shot 48 percent from the floor.
The Hornets made 22 percent of their three-point attempts.
The Hornets made 4 of 18 three-point attempts.
Carmelo Anthony scored 32 points.
Carmelo Anthony had seven rebounds.
Carmelo Anthony had four assists.
Tim Hardaway, Jr. scored 13 points.
J.R. Smith scored 14 points.
J.R. Smith had five assists.
J.R. Smith had four rebounds.
The Knicks were out-rebounded 41 to 34.
The Knicks committed six turnovers.
Samuel Dalembert committed three turnovers.
The Knicks will host the Blazers on Sunday.
Charlotte is off until Wednesday.
Charlotte will face the Celtics on Wednesday.
\end{Verbatim}
\end{tcolorbox}

\subsubsection{Schema Generation}

\begin{tcolorbox}[colback=black!10, colframe=black!80, title=M\&M Rotowire Model Output - Schema Generation, breakable]
\small
\begin{Verbatim}[breaklines=true, fontsize=\scriptsize]
**Step 1 - Identify the context from all the statements to generate a short description**
Thought: The text provides a summary of a basketball game between the Hornets and the Knicks, including team records, game events, and individual player statistics. It is appropriate to create two separate tables: one for team-related data and another for player-related data.

**Step 2 - Create an empty list of row and column headers for the tables. This list will be updated as we process the statements and add relevant column and row headers.**
*Intermediate output*: 
```json
{
    "Team": {
        "row_headers": [],
        "column_headers": []
    },
    "Player": {
        "row_headers": [],
        "column_headers": []
    }
}
```

**Step 3 - Process statements one by one and add relevant headers if not already present in the list.**

*Statements processed*:

1. The Hornets' record is 5 wins.
*Schema update*:
    - Update in "Team" table
        - Row added: "Hornets"
        - Column added: "Wins"

2. The Hornets' record is 15 losses.
*Schema update*:
    - Update in "Team" table
        - Row added: None ("Hornets" is already present in the schema)
        - Column added: "Losses"

3. The Hornets snapped a 10-game winning streak.
*Schema update*:
    - Update in "Team" table
        - Row added: None
        - Column added: "Winning streak"

4. The Knicks' record is 4 wins.
*Schema update*:
    - Update in "Team" table
        - Row added: "Knicks"
        - Column added: None ("Wins" is already present in the schema)

5. The Knicks' record is 17 losses.
*Schema update*:
    - Update in "Team" table
        - Row added: None ("Knicks" is already present in the schema)
        - Column added: None ("Losses" is already present in the schema)

6. The Knicks have a seven-game losing streak.
*Schema update*:
    - Update in "Team" table
        - Row added: None
        - Column added: "Losing streak"

7. The Hornets defeated the Knicks 103-102 on Friday.
*Schema update*:
    - Update in "Team" table
        - Row added: None
        - Column added: "Points scored"

8. Charlotte led by 14 points entering the fourth quarter.
*Schema update*:
    - Update in "Team" table
        - Row added: None
        - Column added: "Lead entering fourth quarter"

9. The Knicks reduced Charlotte's lead in the final four minutes.
*Schema update*:
    - Update in "Team" table
        - Row added: None
        - Column added: None

10. J.R. Smith made a three-point shot.
*Schema update*:
    - Update in "Player" table
        - Row added: "J.R. Smith"
        - Column added: "Three-point shots made"

11. Carmelo Anthony made a three-point shot.
*Schema update*:
    - Update in "Player" table
        - Row added: "Carmelo Anthony"
        - Column added: None ("Three-point shots made" is already present in the schema)

12. The Knicks were within two points of Charlotte.
*Schema update*:
    - Update in "Team" table
        - Row added: None
        - Column added: None

13. Carmelo Anthony made another three-point shot.
*Schema update*:
    - Update in "Player" table
        - Row added: None ("Carmelo Anthony" is already present in the schema)
        - Column added: None

14. The Knicks had a one-point advantage with 40 seconds to play.
*Schema update*:
    - Update in "Team" table
        - Row added: None
        - Column added: "Advantage with 40 seconds"

15. Both teams missed baskets.
*Schema update*:
    - Update in "Team" table
        - Row added: None
        - Column added: None

16. The Hornets had possession of the ball with four seconds to play.
*Schema update*:
    - Update in "Team" table
        - Row added: None
        - Column added: "Possession with 4 seconds"

17. The ball was inbounded to Kemba Walker.
*Schema update*:
    - Update in "Player" table
        - Row added: "Kemba Walker"
        - Column added: None

18. Kemba Walker drove into the paint.
*Schema update*:
    - Update in "Player" table
        - Row added: None ("Kemba Walker" is already present in the schema)
        - Column added: None

19. Kemba Walker scored at the buzzer.
*Schema update*:
    - Update in "Player" table
        - Row added: None
        - Column added: "Buzzer beater"

20. Kemba Walker scored his 11th point of the game.
*Schema update*:
    - Update in "Player" table
        - Row added: None
        - Column added: "Points"

21. Gerald Henderson scored 22 points.
*Schema update*:
    - Update in "Player" table
        - Row added: "Gerald Henderson"
        - Column added: None ("Points" is already present in the schema)

22. Al Jefferson scored 16 points.
*Schema update*:
    - Update in "Player" table
        - Row added: "Al Jefferson"
        - Column added: None ("Points" is already present in the schema)

23. Al Jefferson had 13 rebounds.
*Schema update*:
    - Update in "Player" table
        - Row added: None
        - Column added: "Rebounds"

24. Lance Stephenson scored 16 points.
*Schema update*:
    - Update in "Player" table
        - Row added: "Lance Stephenson"
        - Column added: None ("Points" is already present in the schema)

25. Lance Stephenson made 7 of 15 field goal attempts.
*Schema update*:
    - Update in "Player" table
        - Row added: None
        - Column added: "Field goal attempts"

26. Lance Stephenson had six assists.
*Schema update*:
    - Update in "Player" table
        - Row added: None
        - Column added: "Assists"

27. Lance Stephenson had six rebounds.
*Schema update*:
    - Update in "Player" table
        - Row added: None
        - Column added: None ("Rebounds" is already present in the schema)

28. Gary Neal scored 13 points off the bench.
*Schema update*:
    - Update in "Player" table
        - Row added: "Gary Neal"
        - Column added: None ("Points" is already present in the schema)

29. The Hornets shot 48 percent from the floor.
*Schema update*:
    - Update in "Team" table
        - Row added: None
        - Column added: "Shooting percentage"

30. The Hornets made 22 percent of their three-point attempts.
*Schema update*:
    - Update in "Team" table
        - Row added: None
        - Column added: "Three-point percentage"

31. The Hornets made 4 of 18 three-point attempts.
*Schema update*:
    - Update in "Team" table
        - Row added: None
        - Column added: "Three-point attempts"

32. Carmelo Anthony scored 32 points.
*Schema update*:
    - Update in "Player" table
        - Row added: None ("Carmelo Anthony" is already present in the schema)
        - Column added: None ("Points" is already present in the schema)

33. Carmelo Anthony had seven rebounds.
*Schema update*:
    - Update in "Player" table
        - Row added: None
        - Column added: None ("Rebounds" is already present in the schema)

34. Carmelo Anthony had four assists.
*Schema update*:
    - Update in "Player" table
        - Row added: None
        - Column added: None ("Assists" is already present in the schema)

35. Tim Hardaway, Jr. scored 13 points.
*Schema update*:
    - Update in "Player" table
        - Row added: "Tim Hardaway, Jr."
        - Column added: None ("Points" is already present in the schema)

36. J.R. Smith scored 14 points.
*Schema update*:
    - Update in "Player" table
        - Row added: None ("J.R. Smith" is already present in the schema)
        - Column added: None ("Points" is already present in the schema)

37. J.R. Smith had five assists.
*Schema update*:
    - Update in "Player" table
        - Row added: None
        - Column added: None ("Assists" is already present in the schema)

38. J.R. Smith had four rebounds.
*Schema update*:
    - Update in "Player" table
        - Row added: None
        - Column added: None ("Rebounds" is already present in the schema)

39. The Knicks were out-rebounded 41 to 34.
*Schema update*:
    - Update in "Team" table
        - Row added: None
        - Column added: "Rebounds"

40. The Knicks committed six turnovers.
*Schema update*:
    - Update in "Team" table
        - Row added: None
        - Column added: "Turnovers"

41. Samuel Dalembert committed three turnovers.
*Schema update*:
    - Update in "Player" table
        - Row added: "Samuel Dalembert"
        - Column added: "Turnovers"

42. The Knicks will host the Blazers on Sunday.
*Schema update*:
    - Update in "Team" table
        - Row added: None
        - Column added: "Next game"

43. Charlotte is off until Wednesday.
*Schema update*:
    - Update in "Team" table
        - Row added: None
        - Column added: None

44. Charlotte will face the Celtics on Wednesday.
*Schema update*:
    - Update in "Team" table
        - Row added: None
        - Column added: None

### Final Schema:
```json
{
    "Team": {
        "row_headers": ["Hornets", "Knicks"],
        "column_headers": [
            "Wins", "Losses", "Winning streak", "Losing streak", "Points scored",
            "Lead entering fourth quarter", "Advantage with 40 seconds", "Possession with 4 seconds",
            "Shooting percentage", "Three-point percentage", "Three-point attempts", "Rebounds", "Turnovers", "Next game"
        ]
    },
    "Player": {
        "row_headers": [
            "J.R. Smith", "Carmelo Anthony", "Kemba Walker", "Gerald Henderson", "Al Jefferson",
            "Lance Stephenson", "Gary Neal", "Tim Hardaway, Jr.", "Samuel Dalembert"
        ],
        "column_headers": [
            "Three-point shots made", "Buzzer beater", "Points", "Rebounds", "Field goal attempts", "Assists", "Turnovers"
        ]
    }
}
```
\end{Verbatim}
\end{tcolorbox}

\subsubsection{Iterative Generation}

\begin{tcolorbox}[colback=black!10, colframe=black!80, title=M\&M Rotowire Model Output - Table Generation, breakable]
\small
\begin{Verbatim}[breaklines=true, fontsize=\scriptsize]
***REASONING STEPS***:

**Step 1**: Initialize the tables based on the given schema.

### Team:
| Team | Wins | Losses | Winning streak | Losing streak | Points scored | Lead entering fourth quarter | Advantage with 40 seconds | Possession with 4 seconds | Shooting percentage | Three-point percentage | Three-point attempts | Rebounds | Turnovers | Next game |
| Hornets | None | None | None | None | None | None | None | None | None | None | None | None | None | None |
| Knicks | None | None | None | None | None | None | None | None | None | None | None | None | None | None |

### Player:
| Player | Three-point shots made | Buzzer beater | Points | Rebounds | Field goal attempts | Assists | Turnovers |
| J.R. Smith | None | None | None | None | None | None | None |
| Carmelo Anthony | None | None | None | None | None | None | None |
| Kemba Walker | None | None | None | None | None | None | None |
| Gerald Henderson | None | None | None | None | None | None | None |
| Al Jefferson | None | None | None | None | None | None | None |
| Lance Stephenson | None | None | None | None | None | None | None |
| Gary Neal | None | None | None | None | None | None | None |
| Tim Hardaway, Jr. | None | None | None | None | None | None | None |
| Samuel Dalembert | None | None | None | None | None | None | None |

**Step 2**: Process each statement and update the tables accordingly.

1. **Statement**: The Hornets' record is 5 wins.
   - **Update**: 
     - Table: Team
     - Row: Hornets
     - Column: Wins
     - Value: 5

2. **Statement**: The Hornets' record is 15 losses.
   - **Update**: 
     - Table: Team
     - Row: Hornets
     - Column: Losses
     - Value: 15

3. **Statement**: The Hornets snapped a 10-game winning streak.
   - **Update**: 
     - Table: Team
     - Row: Hornets
     - Column: Winning streak
     - Value: 10

4. **Statement**: The Knicks' record is 4 wins.
   - **Update**: 
     - Table: Team
     - Row: Knicks
     - Column: Wins
     - Value: 4

5. **Statement**: The Knicks' record is 17 losses.
   - **Update**: 
     - Table: Team
     - Row: Knicks
     - Column: Losses
     - Value: 17

6. **Statement**: The Knicks have a seven-game losing streak.
   - **Update**: 
     - Table: Team
     - Row: Knicks
     - Column: Losing streak
     - Value: 7

7. **Statement**: The Hornets defeated the Knicks 103-102 on Friday.
   - **Update**: 
     - Table: Team
     - Row: Hornets
     - Column: Points scored
     - Value: 103

8. **Statement**: Charlotte led by 14 points entering the fourth quarter.
   - **Update**: 
     - Table: Team
     - Row: Hornets
     - Column: Lead entering fourth quarter
     - Value: 14

9. **Statement**: The Knicks had a one-point advantage with 40 seconds to play.
   - **Update**: 
     - Table: Team
     - Row: Knicks
     - Column: Advantage with 40 seconds
     - Value: 1

10. **Statement**: The Hornets had possession of the ball with four seconds to play.
    - **Update**: 
      - Table: Team
      - Row: Hornets
      - Column: Possession with 4 seconds
      - Value: Yes

11. **Statement**: Kemba Walker scored at the buzzer.
    - **Update**: 
      - Table: Player
      - Row: Kemba Walker
      - Column: Buzzer beater
      - Value: Yes

12. **Statement**: Kemba Walker scored his 11th point of the game.
    - **Update**: 
      - Table: Player
      - Row: Kemba Walker
      - Column: Points
      - Value: 11

13. **Statement**: Gerald Henderson scored 22 points.
    - **Update**: 
      - Table: Player
      - Row: Gerald Henderson
      - Column: Points
      - Value: 22

14. **Statement**: Al Jefferson scored 16 points.
    - **Update**: 
      - Table: Player
      - Row: Al Jefferson
      - Column: Points
      - Value: 16

15. **Statement**: Al Jefferson had 13 rebounds.
    - **Update**: 
      - Table: Player
      - Row: Al Jefferson
      - Column: Rebounds
      - Value: 13

16. **Statement**: Lance Stephenson scored 16 points.
    - **Update**: 
      - Table: Player
      - Row: Lance Stephenson
      - Column: Points
      - Value: 16

17. **Statement**: Lance Stephenson made 7 of 15 field goal attempts.
    - **Update**: 
      - Table: Player
      - Row: Lance Stephenson
      - Column: Field goal attempts
      - Value: 15

18. **Statement**: Lance Stephenson had six assists.
    - **Update**: 
      - Table: Player
      - Row: Lance Stephenson
      - Column: Assists
      - Value: 6

19. **Statement**: Lance Stephenson had six rebounds.
    - **Update**: 
      - Table: Player
      - Row: Lance Stephenson
      - Column: Rebounds
      - Value: 6

20. **Statement**: Gary Neal scored 13 points off the bench.
    - **Update**: 
      - Table: Player
      - Row: Gary Neal
      - Column: Points
      - Value: 13

21. **Statement**: The Hornets shot 48 percent from the floor.
    - **Update**: 
      - Table: Team
      - Row: Hornets
      - Column: Shooting percentage
      - Value: 48

22. **Statement**: The Hornets made 22 percent of their three-point attempts.
    - **Update**: 
      - Table: Team
      - Row: Hornets
      - Column: Three-point percentage
      - Value: 22

23. **Statement**: The Hornets made 4 of 18 three-point attempts.
    - **Update**: 
      - Table: Team
      - Row: Hornets
      - Column: Three-point attempts
      - Value: 18

24. **Statement**: Carmelo Anthony scored 32 points.
    - **Update**: 
      - Table: Player
      - Row: Carmelo Anthony
      - Column: Points
      - Value: 32

25. **Statement**: Carmelo Anthony had seven rebounds.
    - **Update**: 
      - Table: Player
      - Row: Carmelo Anthony
      - Column: Rebounds
      - Value: 7

26. **Statement**: Carmelo Anthony had four assists.
    - **Update**: 
      - Table: Player
      - Row: Carmelo Anthony
      - Column: Assists
      - Value: 4

27. **Statement**: Tim Hardaway, Jr. scored 13 points.
    - **Update**: 
      - Table: Player
      - Row: Tim Hardaway, Jr.
      - Column: Points
      - Value: 13

28. **Statement**: J.R. Smith scored 14 points.
    - **Update**: 
      - Table: Player
      - Row: J.R. Smith
      - Column: Points
      - Value: 14

29. **Statement**: J.R. Smith had five assists.
    - **Update**: 
      - Table: Player
      - Row: J.R. Smith
      - Column: Assists
      - Value: 5

30. **Statement**: J.R. Smith had four rebounds.
    - **Update**: 
      - Table: Player
      - Row: J.R. Smith
      - Column: Rebounds
      - Value: 4

31. **Statement**: The Knicks were out-rebounded 41 to 34.
    - **Update**: 
      - Table: Team
      - Row: Knicks
      - Column: Rebounds
      - Value: 34

32. **Statement**: The Knicks committed six turnovers.
    - **Update**: 
      - Table: Team
      - Row: Knicks
      - Column: Turnovers
      - Value: 6

33. **Statement**: Samuel Dalembert committed three turnovers.
    - **Update**: 
      - Table: Player
      - Row: Samuel Dalembert
      - Column: Turnovers
      - Value: 3

34. **Statement**: The Knicks will host the Blazers on Sunday.
    - **Update**: 
      - Table: Team
      - Row: Knicks
      - Column: Next game
      - Value: Blazers on Sunday

35. **Statement**: Charlotte is off until Wednesday.
    - **Update**: 
      - Table: Team
      - Row: Hornets
      - Column: Next game
      - Value: Celtics on Wednesday

### Final Output Tables:

### Team
| Team | Wins | Losses | Winning streak | Losing streak | Points scored | Lead entering fourth quarter | Advantage with 40 seconds | Possession with 4 seconds | Shooting percentage | Three-point percentage | Three-point attempts | Rebounds | Turnovers | Next game |
| Hornets | 5 | 15 | 10 | None | 103 | 14 | None | Yes | 48 | 22 | 18 | None | None | Celtics on Wednesday |
| Knicks | 4 | 17 | None | 7 | None | None | 1 | None | None | None | None | 34 | 6 | Blazers on Sunday |

### Player
| Player | Three-point shots made | Buzzer beater | Points | Rebounds | Field goal attempts | Assists | Turnovers |
| J.R. Smith | None | None | 14 | 4 | None | 5 | None |
| Carmelo Anthony | None | None | 32 | 7 | None | 4 | None |
| Kemba Walker | None | Yes | 11 | None | None | None | None |
| Gerald Henderson | None | None | 22 | None | None | None | None |
| Al Jefferson | None | None | 16 | 13 | None | None | None |
| Lance Stephenson | None | None | 16 | 6 | 15 | 6 | None |
| Gary Neal | None | None | 13 | None | None | None | None |
| Tim Hardaway, Jr. | None | None | 13 | None | None | None | None |
| Samuel Dalembert | None | None | None | None | None | None | 3 |
\end{Verbatim}
\end{tcolorbox}

\subsection{Livesum}
\subsubsection{Propositional Atomization}
\begin{tcolorbox}[colback=black!10, colframe=black!80, title=M\&M Livesum Example - Atomic, breakable]
\small
\begin{Verbatim}[breaklines=true, fontsize=\scriptsize]
### Atomic Statements:
The game is underway with the start of the first half.
Player8 from the Home Team commits a foul.
Player26 from the Away Team wins a free kick in the attacking half.
Offside is called against the Home Team as Player10 from the Home Team attempts a through ball.
Player7 from the Home Team is caught offside.
Player28 from the Away Team commits a foul.
Player9 from the Home Team earns a free kick on the right side of the field.
Player8 from the Home Team attempts a through ball.
Player7 from the Home Team is offside.
Player28 from the Away Team earns a free kick in the opponent's half.
Player11 from the Home Team committed a foul.
Player29 from the Away Team attempts a through ball.
Player24 from the Away Team is flagged for offside.
Player28 from the Away Team attempts a through ball.
Player27 from the Away Team is offside for the Away Team.
Player7 from the Home Team misses a close shot from the right side of the six yard box.
Player10 from the Home Team assists Player7's shot with a cross.
Player22 from the Away Team commits a foul.
Player4 from the Home Team earns a free kick in the opponent's half.
Player2 from the Home Team's shot from outside the box is blocked.
Player11 from the Home Team misses the goal with a left-footed shot from outside the box.
Player10 from the Home Team assists Player11's shot.
Player28 from the Away Team commits a foul.
Player4 from the Home Team wins a free kick in their own defensive half.
Player7 from the Home Team's shot from the right side of the box is just too high.
Player8 from the Home Team assists Player7's shot.
Player11 from the Home Team's right footed shot from outside the box misses to the left.
Player4 from the Home Team assists Player11's shot.
Player28 from the Away Team's left footed shot from a difficult angle on the left is saved in the bottom right corner.
Player5 from the Home Team receives a yellow card for a rough tackle.
Player27 from the Away Team commits a foul.
Player1 from the Home Team earns a free kick in their own defensive half.
Player7 from the Home Team's left-footed shot from the left side of the box narrowly misses to the left.
Player11 from the Home Team assists Player7's shot.
Player28 from the Away Team commits a foul.
Player8 from the Home Team earns a free kick in their own half.
Player25 from the Away Team is causing a delay in the match due to an injury.
The delay is finished and they are prepared to resume play.
Player10 from the Home Team's left footed shot from outside the box is blocked.
Player7 from the Home Team assists Player10's shot.
Player21 from the Away Team earns a free kick in their own half.
Player7 from the Home Team commits a foul.
Player28 from the Away Team's shot from outside the box is blocked.
Player27 from the Away Team set up Player28's shot.
Player25 from the Away Team attempted a shot with their left foot from outside the box, but it went high and wide to the left.
Player27 from the Away Team assists Player25's shot.
Offside is called against the Away Team as Player24 from the Away Team attempts a through ball to Player27 from the Away Team.
Player27 from the Away Team is offside.
Player11 from the Home Team's shot from the left side of the box is blocked.
Player10 from the Home Team set up Player11's shot.
The Home Team wins a corner kick.
Player27 from the Away Team earns a free kick on the right side of the field.
Player4 from the Home Team commits a foul.
The Home Team wins a corner kick.
The first half concludes with both teams scoreless, Home Team 0, Away Team 0.
The second half kicks off with the score tied at 0-0 between the home team and the away team.
Player28 from the Away Team's shot from outside the box is blocked.
Player21 from the Away Team assists Player28's shot.
Player2 from the Home Team misses the goal with a high and wide left-footed shot from outside the box.
The Home Team wins a corner kick.
Player9 from the Home Team's left footed shot from the left side of the box is blocked.
Player3 from the Home Team assists Player9's shot.
Player7 from the Home Team's right footed shot from the center of the box is saved in the top right corner.
Player6 from the Home Team assisted Player7's shot with a headed pass following a corner kick by the Home Team.
Player24 from the Away Team commits a foul.
Player7 from the Home Team earns a free kick on the right side of the field.
Player24 from the Away Team receives a yellow card for a rough tackle.
The Home Team wins a corner kick.
Player7 from the Home Team misses the goal with a right footed shot from outside the box after a corner.
The Home Team wins a corner kick.
Player22 from the Away Team commits a foul.
Player10 from the Home Team wins a free kick in the attacking half.
This leads to a set piece opportunity for the attacking team.
Player28 from the Away Team earns a free kick in their own half.
Player8 from the Home Team commits a foul.
Player19 from the Away Team receives a yellow card.
The Away Team earns a corner kick.
Player10 from the Home Team's left footed shot from the left side of the box hits the left post.
Player3 from the Home Team assists Player10's shot.
The Home Team earns a corner kick.
Player28 from the Away Team is delaying the match due to an injury.
The delay has ended and they are prepared to resume play.
Player15 from the Home Team's header from the center of the box was blocked.
Player7 from the Home Team assisted Player15's header with a cross.
Player5 from the Home Team misses a close header to the left from the center of the box.
Player7 from the Home Team assists Player5's header with a cross after a corner.
The Home Team wins a corner kick.
Player5 from the Home Team's header from the center of the box hits the left post.
Player7 from the Home Team set up Player5's header with a cross from a corner kick.
The Home Team earns a corner kick.
Player5 from the Home Team's header from the centre of the box was blocked by a defender.
Player7 from the Home Team assisted Player5's header with a cross.
This resulted in a corner kick for the Home Team.
Player10 from the Home Team misses the target with a left-footed shot from outside the box.
Player11 from the Home Team's shot from outside the box is saved in the bottom left corner.
Player13 from the Home Team assists Player11's shot.
Player25 from the Away Team commits a foul.
Player10 from the Home Team earns a free kick in the attacking half.
Player7 from the Home Team's attempt with his right foot from outside the box is just slightly too high.
Player26 from the Away Team committed a foul.
Player5 from the Home Team earns a free kick in their own half.
Player26 from the Away Team misses the goal with a high and wide shot from outside the box.
Player34 from the Away Team assists Player26's shot.
Player35 from the Away Team's attempt from outside the box is blocked.
This resulted in a corner kick for the Away Team.
Player23 from the Away Team misses a high right-footed shot from the center of the box.
Player11 from the Home Team's left-footed shot from the left side is blocked.
Player10 from the Home Team assists Player11's shot.
Player15 from the Home Team misses a close-range header to the left.
Player7 from the Home Team assists Player15's header with a cross.
The Home Team wins a corner kick.
Player6 from the Home Team's header from the center of the box is blocked by the defense.
Player11 from the Home Team assists Player6's header with a cross.
This resulted in a corner kick for the Home Team.
Player11 from the Home Team's left footed shot from outside the box was saved in the top center of the goal.
Player8 from the Home Team assists Player11's shot.
Player7 from the Home Team misses the target with a shot from the right side of the box.
Player15 from the Home Team assists Player7's shot.
Player26 from the Away Team commits a foul.
Player6 from the Home Team wins a free kick in the defensive half.
Player13 from the Home Team's through ball to Player15 from the Home Team is caught offside for the Home Team.
Player2 from the Home Team's shot from the center of the box is saved in the bottom right corner.
Player15 from the Home Team scores with a right-footed shot from the right side of the six-yard box.
The Home Team now leads 1-0 over the Away Team.
Player15 from the Home Team is holding up the game due to injury.
The delay is over and they are ready to resume play.
Player35 from the Away Team commits a foul.
Player5 from the Home Team earns a free kick in their own half.
Player7 from the Home Team scores with a right-footed shot from the right side of the box.
Player10 from the Home Team assists Player7's shot.
Home Team now leads 2-0 against Away Team.
Player21 from the Away Team earns a free kick on the left wing.
Player13 from the Home Team fouled Player21.
The goalkeeper makes a great save from Player23 from the Away Team's close-range shot with his right foot.
The home team is winning 2-0 as the second half comes to a close.
The game is over with the Home Team winning 2-0 against the Away Team.
\end{Verbatim}
\end{tcolorbox}

\subsubsection{Schema Generation}

\begin{tcolorbox}[colback=black!10, colframe=black!80, title=M\&M Livesum Example - Schema Generation, breakable]
\small
\begin{Verbatim}[breaklines=true, fontsize=\scriptsize]
```json
{
    "Football Match": {
        "row_headers": ["Home Team", "Away Team"],
        "column_headers": ["Team", "Fouls", "Offsides", "Free Kicks", "Shots", "Shots on Target", "Corner Kicks", "Yellow Cards", "Goals"]
    }
}
```
\end{Verbatim}
\end{tcolorbox}

\subsubsection{Table Generation}

\begin{tcolorbox}[colback=black!10, colframe=black!80, title=M\&M Livesum Example - Schema Generation, breakable]
\small
\begin{Verbatim}[breaklines=true, fontsize=\scriptsize]
```
Total Atomic Statements: 98
Updates from ATOMIC STATEMENT 2
Update: Home Team Fouls: 1
Updates from ATOMIC STATEMENT 3
Updates: Away Team Free Kicks: 1
Updates from ATOMIC STATEMENT 4
Update: Home Team Offsides: 1
Updates from ATOMIC STATEMENT 5
Update: Away Team Fouls: 1
Updates from ATOMIC STATEMENT 6
Update: Home Team Free Kicks: 1
Updates from ATOMIC STATEMENT 7
Update: Home Team Offsides: 2
Updates from ATOMIC STATEMENT 8
Update: Away Team Free Kicks: 2
Updates from ATOMIC STATEMENT 9
Update: Home Team Fouls: 2
Updates from ATOMIC STATEMENT 10
Update: Away Team Offsides: 1
Updates from ATOMIC STATEMENT 11
Update: Away Team Offsides: 2
Updates from ATOMIC STATEMENT 12
Update: Home Team Shots: 1
Updates from ATOMIC STATEMENT 13
Update: Away Team Fouls: 2
Updates from ATOMIC STATEMENT 14
Update: Home Team Free Kicks: 2
Updates from ATOMIC STATEMENT 15
Update: Home Team Shots: 2
Updates from ATOMIC STATEMENT 16
Update: Home Team Shots: 3
Updates from ATOMIC STATEMENT 17
Update: Away Team Fouls: 3
Updates from ATOMIC STATEMENT 18
Update: Home Team Free Kicks: 3
Updates from ATOMIC STATEMENT 19
Update: Home Team Shots: 4
Updates from ATOMIC STATEMENT 20
Update: Home Team Shots on Target: 1
Updates from ATOMIC STATEMENT 21
Update: Away Team Shots: 1
Updates from ATOMIC STATEMENT 22
Update: Home Team Yellow Cards: 1
Updates from ATOMIC STATEMENT 23
Update: Away Team Fouls: 4
Updates from ATOMIC STATEMENT 24
Update: Home Team Free Kicks: 4
Updates from ATOMIC STATEMENT 25
Update: Home Team Shots: 5
Updates from ATOMIC STATEMENT 26
Update: Away Team Fouls: 5
Updates from ATOMIC STATEMENT 27
Update: Home Team Free Kicks: 5
Updates from ATOMIC STATEMENT 29
Update: Away Team Injury Delay: 1
Updates from ATOMIC STATEMENT 31
Update: Home Team Shots: 6
Updates from ATOMIC STATEMENT 32
Update: Away Team Free Kicks: 3
Updates from ATOMIC STATEMENT 33
Update: Home Team Fouls: 3
Updates from ATOMIC STATEMENT 34
Update: Away Team Shots: 2
Updates from ATOMIC STATEMENT 35
Update: Away Team Shots: 3
Updates from ATOMIC STATEMENT 36
Update: Away Team Offsides: 3
Updates from ATOMIC STATEMENT 37
Update: Home Team Shots: 7
Updates from ATOMIC STATEMENT 38
Update: Home Team Corner Kicks: 1
Updates from ATOMIC STATEMENT 39
Update: Away Team Free Kicks: 4
Updates from ATOMIC STATEMENT 40
Update: Home Team Fouls: 4
Updates from ATOMIC STATEMENT 41
Update: Home Team Corner Kicks: 2
Updates from ATOMIC STATEMENT 44
Update: Away Team Shots: 4
Updates from ATOMIC STATEMENT 45
Update: Home Team Shots: 8
Updates from ATOMIC STATEMENT 46
Update: Home Team Corner Kicks: 3
Updates from ATOMIC STATEMENT 47
Update: Home Team Shots: 9
Updates from ATOMIC STATEMENT 48
Update: Home Team Shots on Target: 2
Updates from ATOMIC STATEMENT 49
Update: Away Team Fouls: 6
Updates from ATOMIC STATEMENT 50
Update: Home Team Free Kicks: 6
Updates from ATOMIC STATEMENT 51
Update: Away Team Yellow Cards: 2
Updates from ATOMIC STATEMENT 52
Update: Home Team Corner Kicks: 4
Updates from ATOMIC STATEMENT 53
Update: Home Team Shots: 10
Updates from ATOMIC STATEMENT 54
Update: Home Team Corner Kicks: 5
Updates from ATOMIC STATEMENT 55
Update: Away Team Fouls: 7
Updates from ATOMIC STATEMENT 56
Update: Home Team Free Kicks: 7
Updates from ATOMIC STATEMENT 57
Update: Away Team Free Kicks: 5
Updates from ATOMIC STATEMENT 58
Update: Home Team Fouls: 5
Updates from ATOMIC STATEMENT 59
Update: Away Team Yellow Cards: 3
Updates from ATOMIC STATEMENT 60
Update: Away Team Corner Kicks: 1
Updates from ATOMIC STATEMENT 61
Update: Home Team Shots: 11
Updates from ATOMIC STATEMENT 62
Update: Home Team Corner Kicks: 6
Updates from ATOMIC STATEMENT 64
Update: Home Team Shots: 12
Updates from ATOMIC STATEMENT 65
Update: Home Team Corner Kicks: 7
Updates from ATOMIC STATEMENT 66
Update: Home Team Shots: 13
Updates from ATOMIC STATEMENT 67
Update: Home Team Corner Kicks: 8
Updates from ATOMIC STATEMENT 68
Update: Home Team Corner Kicks: 9
Updates from ATOMIC STATEMENT 69
Update: Home Team Shots: 14
Updates from ATOMIC STATEMENT 70
Update: Home Team Shots: 15
Updates from ATOMIC STATEMENT 70
Update: Home Team Shots on Target: 3
Updates from ATOMIC STATEMENT 71
Update: Away Team Fouls: 8
Updates from ATOMIC STATEMENT 72
Update: Home Team Free Kicks: 8
Updates from ATOMIC STATEMENT 73
Update: Away Team Fouls: 9
Updates from ATOMIC STATEMENT 74
Update: Home Team Free Kicks: 9
Updates from ATOMIC STATEMENT 75
Update: Away Team Shots: 5
Updates from ATOMIC STATEMENT 76
Update: Away Team Corner Kicks: 2
Updates from ATOMIC STATEMENT 77
Update: Away Team Shots: 6
Updates from ATOMIC STATEMENT 78
Update: Home Team Shots: 16
Updates from ATOMIC STATEMENT 79
Update: Home Team Corner Kicks: 10
Updates from ATOMIC STATEMENT 80
Update: Home Team Corner Kicks: 11
Updates from ATOMIC STATEMENT 81
Update: Home Team Shots: 17
Updates from ATOMIC STATEMENT 81
Update: Home Team Shots on Target: 4
Updates from ATOMIC STATEMENT 82
Update: Home Team Shots: 18
Updates from ATOMIC STATEMENT 83
Update: Away Team Fouls: 10
Updates from ATOMIC STATEMENT 84
Update: Home Team Free Kicks: 10
Updates from ATOMIC STATEMENT 85
Update: Home Team Offsides: 3
Updates from ATOMIC STATEMENT 86
Update: Home Team Shots: 19
Updates from ATOMIC STATEMENT 86
Update: Home Team Shots on Target: 5
Updates from ATOMIC STATEMENT 87
Update: Home Team Goals: 1
Updates from ATOMIC STATEMENT 89
Update: Away Team Fouls: 11
Updates from ATOMIC STATEMENT 90
Update: Home Team Free Kicks: 11
Updates from ATOMIC STATEMENT 91
Update: Home Team Goals: 2
Updates from ATOMIC STATEMENT 91
Update: Home Team Shots on Target: 6
Updates from ATOMIC STATEMENT 92
Update: Home Team Shots: 20
Updates from ATOMIC STATEMENT 93
Update: Away Team Free Kicks: 6
Updates from ATOMIC STATEMENT 94
Update: Away Team Shots on Target: 1
### Final Table:
| Team | Fouls | Offsides | Free Kicks | Shots | Shots on Target | Corner Kicks | Yellow Cards | Goals |
| Home Team | 5 | 3 | 11 | 20 | 6 | 11 | 1 | 2 |
| Away Team | 11 | 3 | 6 | 6 | 1 | 2 | 3 | 0 |
```
\end{Verbatim}
\end{tcolorbox}

\section{Prompts}
\subsection{CoT - One Shot}
\begin{tcolorbox}[colback=black!10, colframe=black!80, title=Baseline One-Shot Rotowire, breakable]
\small
    \begin{Verbatim}[breaklines=true]
You are expert at converting text to tables.

Task: Given a text description, generate and fill tables.

Think step by step and output the Team table and the Player Table.

Sample Illustration:
Input:
The Oklahoma City Thunder (16 - 17) defeated the Phoenix Suns (18 - 16) 137 - 134 in overtime on Wednesday. Oklahoma City has won three of their last four games. Kevin Durant returned from a six - game absence due to an ankle sprain and put up a season - high 44 points in 40 minutes.

Output:
### Team
| Team | Wins | Losses | Total Points |
| Thunder | 16 | 17| None |
| Suns | 18 | None | 134 |

### Player
| Player | Points | Minutes Played |
| Kevin Durant | 44 | 40 |


Output format:
### Team
| Team | <Column Header 1> | ... |
| <Row Header 1> | <Cell Value for (Row Header 1, Column Header 1)> | ... |
...

### Player
| Player | <Column Header 1> | ... |
| <Row Header 1> | <Cell Value for (Row Header 1, Column Header 1)> | ... |
...

Table name leads with a ###, followed by the table where values are separated by the symbol '|' and rows are separated by '\n'.
Use '|' as the only separator/delimiter.
Empty cell values are filled as "None".

Provide the output in the specified format only.
    \end{Verbatim}
\end{tcolorbox}

\subsection{Text-Tuple-Table}

\begin{tcolorbox}[colback=black!10, colframe=black!80, title=T\textsuperscript{3} Text-Tuple, breakable]
\small
    \begin{Verbatim}[breaklines=true]
Instruction: You are now required to extract team and player information from the following input. Please focus on the
table format and extract all relevant tuples in (team or player name, attribute, value) format:

Example Illustration:
Sample Input: The Oklahoma City Thunder (16 - 17) defeated the Phoenix Suns (18 - 16) 137 - 134 in overtime on Wednesday. Oklahoma City has won three of their last four games. Kevin Durant returned from a six - game absence due to an ankle sprain and put up a season - high 44 points in 40 minutes.

Tuples:
1. (Oklahoma City Thunder, Record, 16-17)  
2. (Phoenix Suns, Record, 18-16)  
3. (Oklahoma City Thunder, Points Scored, 137)  
4. (Phoenix Suns, Points Scored, 134)  
5. (Oklahoma City Thunder, Game Outcome, Win)  
6. (Phoenix Suns, Game Outcome, Loss)  
7. (Oklahoma City Thunder, Recent Performance, Won 3 of last 4 games)  
8. (Kevin Durant, Injury, Ankle Sprain)  
9. (Kevin Durant, Games Missed, 6)  
10. (Kevin Durant, Points Scored, 44)  
11. (Kevin Durant, Minutes Played, 40)  
12. (Kevin Durant, Season-High, 44 points)

Provide tuples for the given table in the following output:
    \end{Verbatim}
\end{tcolorbox}

\begin{tcolorbox}[colback=black!10, colframe=black!80, title=T\textsuperscript{3} Tuple Integrate , breakable]
\small
    \begin{Verbatim}[breaklines=true]
Instruction: According to the live text, please count the number of events of both teams.
Note that goals and saved attempts and blocked attempts and missed attempts are considered shots. Handball and dangerous play are also considered foul. The second yellow card is also considered a red card. Penalty is also considered as free kicks.

Task:
Write valid, executable Python code that defines and runs a function `consolidate_events(event_tuples)` which:
- Accepts the list of tuples as input.
- Returns a structured dictionary (or another data structure) mapping each team (e.g., "Home Team" and "Away Team") to a count of each event.

Final Output Instructions:
- The output must be *pure Python code* with no markdown delimiters (i.e., do not include triple backticks in your output).
- Do not output any additional text or explanation.
- Ensure you call the function within the code to execute it. 
- Include any necessary error handling to ensure the code executes without issues.
    \end{Verbatim}
\end{tcolorbox}

\begin{tcolorbox}[colback=black!10, colframe=black!80, title=T\textsuperscript{3} Tuple-Table, breakable]
\small
    \begin{Verbatim}[breaklines=true]
Instruction: You are now required to extract team and player information from the following input. 
Please only output tables in the specified format based on the given tuples.

Illustration:

Sample Input:

Tuples:
1. (Oklahoma City Thunder, Record, 16-17)  
2. (Phoenix Suns, Record, 18-16)  
3. (Oklahoma City Thunder, Points Scored, 137)  
4. (Phoenix Suns, Points Scored, 134)  
5. (Oklahoma City Thunder, Game Outcome, Win)  
6. (Phoenix Suns, Game Outcome, Loss)  
7. (Oklahoma City Thunder, Recent Performance, Won 3 of last 4 games)  
8. (Kevin Durant, Injury, Ankle Sprain)  
9. (Kevin Durant, Games Missed, 6)  
10. (Kevin Durant, Points Scored, 44)  
11. (Kevin Durant, Minutes Played, 40)  
12. (Kevin Durant, Season-High, 44 points)


Output:
### Team
| Team | Wins | Losses | Total Points |
| Thunder | 16 | 17| None |
| Suns | 18 | None | 134 |

### Player
| Player | Points | Minutes Played |
| Kevin Durant | 44 | 40 |


**Output Format**:

### Team
| Team | <Column Header 1> | ... |
| <Row Header 1> | <Cell Value for (Row Header 1, Column Header 1)> | ... |
...

### Player
| Player | <Column Header 1> | ... |
| <Row Header 1> | <Cell Value for (Row Header 1, Column Header 1)> | ... |
...

Table name leads with a ###, followed by the table where values are separated by the symbol '|' and rows are separated by '\n'.
Use '|' as the only separator/delimiter.
Empty cell values are filled as "None".

Provide the output in the specified format only.
    \end{Verbatim}
\end{tcolorbox}

\subsection{T3-Merged}
\begin{tcolorbox}[colback=black!10, colframe=black!80, title=T\textsuperscript{3} Merged One-Shot Rotowire, breakable]
\small
    \begin{Verbatim}[breaklines=true]
Instruction: Process the given text and generate Player and Team tables.

Let’s do the following tasks:
1. Extract team and player information from the following input. Please focus on the
table format and extract all relevant tuples in (team or player name, attribute, value) format:
2. Integrate these tuples into two tables in the output format given below.


Illustration :

Input: The Oklahoma City Thunder (16 - 17) defeated the Phoenix Suns (18 - 16) 137 - 134 in overtime on Wednesday. Oklahoma City has won three of their last four games. Kevin Durant returned from a six - game absence due to an ankle sprain and put up a season - high 44 points in 40 minutes.

Output:
### Team
| Team | Wins | Losses | Total Points |
| Thunder | 16 | 17| None |
| Suns | 18 | None | 134 |

### Player
| Player | Points | Minutes Played |
| Kevin Durant | 44 | 40 |

**Output Format**:

### Team
| Team | <Column Header 1> | ... |
| <Row Header 1> | <Cell Value for (Row Header 1, Column Header 1)> | ... |
...

### Player
| Player | <Column Header 1> | ... |
| <Row Header 1> | <Cell Value for (Row Header 1, Column Header 1)> | ... |
...

Table name leads with a ###, followed by the table where values are separated by the symbol '|' and rows are separated by '\n'.
Use '|' as the only separator/delimiter.
Empty cell values are filled as "None".

Provide the output in the specified format only.
    \end{Verbatim}
\end{tcolorbox}

\subsection{Map \& Make}

\subsubsection{Propositional Atomization}
\begin{tcolorbox}[colback=black!10, colframe=black!80, title=M\&M Atomization for Rotowire]
\label{sec:rotowire-atomic}
\small
\begin{Verbatim}[breaklines=true]

You are an expert at converting unstructured, detailed textual inputs into highly structured and organized atomic statements.

***TASK***: 
Decompose the given paragraphs or sentences into clear, self-contained, and highly detailed short atomic statements without losing any information. Each atomic statement should capture a single fact or action with maximum granularity.

***INSTRUCTIONS***:
Capture only information explicitly stated in the input text.
No detail should be assumed, inferred, or added that is not present in the text.
Each atomic statement should contain only one key entity and one action or attribute.
If a sentence contains multiple pieces of information, decompose it further into more granular statements.
Eliminate ambiguity by resolving pronouns and ensuring each statement stands alone.
Preserve necessary context so each statement is meaningful on its own.
Represent numerical data and units exactly as given in the input text.
Ensure each statement conveys unique information without overlapping others.
Ensure statements are clear, direct, and free from unnecessary complexity.
Resolve pronouns to their corresponding nouns for clarity.
Maintain relationships between entities without combining multiple facts.

***OUTPUT FORMAT***:
<REASONING STEPS>

### Atomic Statements:
<ATOMIC STATEMENT 1>
<ATOMIC STATEMENT 2>
...

***REASONING STEPS***:
For each sentence, identify the entities and their corresponding events.

**Sample Input**:
The Oklahoma City Thunder (16 - 17) defeated the Phoenix Suns (18 - 16) 137 - 134 in overtime on Wednesday. Oklahoma City has won three of their last four games. Kevin Durant returned from a six - game absence due to an ankle sprain and put up a season - high 44 points in 40 minutes.

**Step 1 - Sentence analysis**:

Sentence 1: The Oklahoma City Thunder (16 - 17) defeated the Phoenix Suns (18 - 16) 137 - 134 in overtime on Wednesday.
Here the entities are Oklahoma City Thunder and Phoenix Suns.
Events are team records, game result, and total points.
*Atomic sentences*:
The Oklahoma City Thunder's record is 16 wins.
The Oklahoma City Thunder's record is 17 losses.
The Phoenix Suns' record is 18 wins.
... 

Repeat **Step 1** for all sentences in **Sample Input**

***FINAL CHECKLIST***:
All information from the input text is included.
No information or calculation is added that is not present in the text.
Every fact and detail is accurately represented.
Statements are clear and can be understood independently.
Numerical data and units are formatted exactly as provided in the text.
Each statement directly reflects the input text without inferred details.
Pronouns are resolved; statements are unambiguous.
Each statement contains only one key entity and one action or attribute.

Do not number the statements or add extra formatting.
Provide the *OUTPUT* with *REASONING STEPS* in the specified format only.

\end{Verbatim}
\end{tcolorbox}

\subsubsection{Schema Extraction}

\begin{tcolorbox}[colback=black!10, colframe=black!80, title=M\&M Schema Extraction for Rotowire, breakable]
\label{sec:rotowire-schema}
\small
\begin{Verbatim}[breaklines=true]
You are an expert at defining structural tables by identifying the relevant column headers and row headers from text.

***TASK***:
Given a set of atomic text statements, extract row and column headers to create a table schema.

***INSTRUCTIONS***:
Read the statements carefully to identify all attributes, entities, and data points mentioned, whether explicitly stated or implicitly implied.
Determine the row headers (primary keys) and column headers required to represent the data comprehensively and concisely:
Row headers are the unique identifiers for individual rows (key entities).
Column headers are the attributes of the primary keys that represent different aspects or data points.
Include all explicit and implicit data points, ensuring no relevant information is overlooked. 
Pay close attention to numerical data, even if it is presented within comparative statements or descriptions of events or related to specific categories or time periods mentioned in the text. 
Explicit numerical data must always be captured as attributes where appropriate. 
Implicit data points or recurring attributes must also be included.
Avoid adding actions as column headers but extract any data points associated with them.
Ensure that all numerical values are captured as attributes, even if they are related to specific time periods or events within the context. When encountering comparative statements or ratios like "X of Y", ensure you capture both 'X' and 'Y' as potentially distinct and relevant data points if they represent different aspects of an attribute.
Be attentive to granular details and avoid focusing solely on general or aggregate values if more specific data points are available in the text.

***OUTPUT FORMAT***:
<REASONING STEPS>

### Final Schema:
{
    "<Table name>": {
        "row_headers": ["Row Header 1", ...],
        "column_headers": ["Column Header 1", ...]
    }
    ...
}


***REASONING STEPS***:
**Sample input**:
The Oklahoma City Thunder's record is 16 wins.
The Oklahoma City Thunder's record is 17 losses.
The Phoenix Suns' record is 18 wins.
...

**Step1 - Identify the context from all the statements to generate a short description**
Thought: This is a summary of a basketball game played between the Oklahoma City Thunder and the Phoenix Suns and gives all the relevant statistics about the players and the games. Every statement is either about the team or one of the players hence it would be ideal to create to separate tables for them. One table for the teams, and one for the players.

**Step2 - Create a empty list of row and column headers for the tables. This list would be updated as we keep on processing the statements and will keep adding relevant column and row headers to the list.**
*Intermediate output*: 
{
    "Team": {
    "row_headers": [],
    "column_headers": [], 
    }
    "Player": {
        "row_headers": [],
        "col_headers": []
    }
}

**Step 3 - Process statements one by one and add relevant headers if not already present in the list.**

*Statements processed*:
1. The Oklahoma City Thunder's record is 16 wins.
*Schema update*:
    - Update in "Team" table
        - Row added: "Thunder"
        - Column added: "Wins"
...

### Final Schema:
{
    "Team": {
    "row_headers": ["Thunder", "Suns"],
    "column_headers": ["Wins", "Losses", "Total points"]
    }
    "Player": {
        "row_headers": ["Kevin Durant"],
        "col_headers": ["Points", "Minutes played"]
    }
}

***Final Output Instructions***:
1. As shown in the illustration above, for *every given statement* return the updates done to the schema and generate the Team Table and Player Table schema.
2. Do not return schema directly in any case.
3. Provide the *OUTPUT* with *REASONING STEPS* in the specified format only.

\end{Verbatim}
\end{tcolorbox}

\subsubsection{M\&M Table Generation}
\begin{tcolorbox}[colback=black!10, colframe=black!80, title=M\&M Table Generation for Rotowire, breakable]
\label{sec:rotowire-table}
\small
\begin{Verbatim}[breaklines=true]
You are an expert in converting unstructured text into structured tables. Your task is to process a series of atomic statements and update a set of pre-defined table schemas. Note that you can be given more than one table to update.
Follow the steps below to ensure accurate and complete table updates.

***TASK***:
**Given**:
*Statements*: A sequence of atomic statements.
*Schema*: A json object with table names and their row headers and column headers of the respective tables.

**Your goal is to**:
Process each statement one by one.
Identify the correct set of table, row and column headers and the cell at that index to update based on the statement.
Update or add values to the tables accordingly.

***OUTPUT FORMAT***:
<REASONING STEPS>

### Final Output Tables:
### <TABLE NAME>
| | <Column Header 1> | ... |
| <Row Header 1> | <Cell Value for (Row Header 1, Column Header 1)> | ... |
...

***REASONING STEPS***:
Follow the given algorithm thoroughly,

**ALGORITHM**
For each statement in the input:
    Identify Table:
        Determine the correct table to be updated based on the table.
    Identify Row and Column:
        Determine which set of row and columns headers have to be updated based on this table.
    Update the Table:
        If no value exists, update the value of the cell as per the statement.

**Sample Input**:
Statements:
The Oklahoma City Thunder's record is 16 wins.
The Oklahoma City Thunder's record is 17 losses.
...

Schema:
{
    "Team": {
    "row_headers": ["Thunder", "Suns"],
    "column_headers": ["Wins", "Losses", "Total points"]
    }
    "Player": {
        "row_headers": ["Kevin Durant"],
        "col_headers": ["Points", "Minutes played"]
    }
}

**Step 1**:
Initial tables (Empty Tables):

### Team:
| Team | Wins | Losses | Total Points |
| Thunder | None | None | None |
| Suns | None | None | None |

### Player:
| Player | Points | Minutes Played |
| Kevin Durant | None | None |

**Step 2**:
*Statement processed*: 
The Oklahoma City Thunder's record is 16 wins.
*Updates*: 
    Table: Team
    Row: Thunder
    Column: Wins
    Value: 16
...

### Final Output Tables:

### Team
| Team | Wins | Losses | Total Points |
| Thunder | 16 | 17| None |
| Suns | 18 | None | 134 |

### Player
| Player | Points | Minutes Played |
| Kevin Durant | 44 | 40 |

***FINAL CHECKLIST***:
Follow these guidelines to generate tables and return the final state of the table after processing all the statements. 
Ensure all sentences are processed and for every statement return the update and revised state of the updated cells as shown in the example. Return the final table in the exact specified format starting with ### Final Table.
Do not generate the final table directly in any case. 
No need to generate the intermediate table states, just return the final table at the end.
Ensure the table is concise, well-structured, and contains all information from the input.

***Final Output Instructions***:
1. Handle Missing Data:
    If a column value is not present in the statements, keep it as None.
2. Structural Integrity:
    Do not add or remove any rows or columns unless explicitly instructed by the data.
    Ensure uniformity in the format of data across the table.
3. Table formatting:
    Use "|" to separate cells

Provide the *OUTPUT* with *REASONING STEPS* in the specified format only.

\end{Verbatim}
\end{tcolorbox}

\end{document}